%% file: main.tex
\documentclass[lettersize,journal]{IEEEtran}
\usepackage{amsmath,amsfonts}
\usepackage{algorithm}
\usepackage{algorithmic}
\usepackage{array}
\usepackage{textcomp}
\usepackage{stfloats}
\usepackage{url}
\usepackage{verbatim}
\usepackage{graphicx}
\usepackage{enumitem}
\usepackage{multirow}
\usepackage{threeparttable}
\usepackage{cite}
\usepackage{subfigure}
\usepackage{booktabs}

\input{math_commands.tex}

\usepackage{float}
\newfloat{figtab}{htb}{fgtb}
\makeatletter
\newcommand\figcaption{\def\@captype{figure}\caption}
\newcommand\tabcaption{\def\@captype{table}\caption}

\usepackage{scalerel}
\usepackage{tikz}
\usetikzlibrary{svg.path}
\definecolor{orcidlogocol}{HTML}{A6CE39}
\tikzset{
    orcidlogo/.pic={
        \fill[orcidlogocol] svg{M256,128c0,70.7-57.3,128-128,128C57.3,256,0,198.7,0,128C0,57.3,57.3,0,128,0C198.7,0,256,57.3,256,128z};
        \fill[white] svg{M86.3,186.2H70.9V79.1h15.4v48.4V186.2z}
        svg{M108.9,79.1h41.6c39.6,0,57,28.3,57,53.6c0,27.5-21.5,53.6-56.8,53.6h-41.8V79.1z M124.3,172.4h24.5c34.9,0,42.9-26.5,42.9-39.7c0-21.5-13.7-39.7-43.7-39.7h-23.7V172.4z}
        svg{M88.7,56.8c0,5.5-4.5,10.1-10.1,10.1c-5.6,0-10.1-4.6-10.1-10.1c0-5.6,4.5-10.1,10.1-10.1C84.2,46.7,88.7,51.3,88.7,56.8z};
    }
}

\newcommand\orcidicon[1]{\href{https://orcid.org/#1}{\mbox{\scalerel*{
                \begin{tikzpicture}[yscale=-1,transform shape]
                \pic{orcidlogo};
                \end{tikzpicture}
            }{|}}}}
\usepackage[implicit=false]{hyperref}
\hypersetup{hidelinks,
	colorlinks=true,
	allcolors=black,
	pdfstartview=Fit,
	breaklinks=true}
	
\usepackage{xcolor}
\newcommand{\Revise}[1]{\textcolor{black}{#1}}
\newcommand{\Revisegreen}[1]{\textcolor{black}{#1}}

\def\BibTeX{{\rm B\kern-.05em{\sc i\kern-.025em b}\kern-.08em
    T\kern-.1667em\lower.7ex\hbox{E}\kern-.125emX}}
\usepackage{balance}

\begin{document}
\title{Adaptive Channel Allocation for Robust Differentiable Architecture Search}

\author{Chao Li, Jia Ning, Han Hu, 
        Kun He
\thanks{Chao Li was with the School of Computer Science, Huazhong University of Science and Technology, Wuhan, China 430074.
He is now with the China Mobile Information Technology Co.,Ltd, 
Guangzhou, China 518000. (E-mail: d201880880@hust.edu.cn)}

\thanks{Jia Ning and Kun He are with the School of Computer Science, Huazhong University of Science and Technology, Wuhan, China 430074. (E-mail: \{ninja, brooklet60\}@hust.edu.cn)}

\thanks{Han Hu is with Microsoft Research Asia, Beijing, China 100080. (E-mail: hanhu@microsoft.com)}

%
%
}

\markboth{
}%
{Adaptive Channel Allocation for Robust Differentiable Architecture Search}

\maketitle

\begin{abstract}
Differentiable ARchiTecture Search (DARTS) has attracted much attention due to its simplicity and significant improvement in efficiency. However, the excessive accumulation of the \textit{skip connection}, when training epochs become large, makes it suffer from weak stability and low robustness, thus limiting its practical applications. Many works have attempted to restrict the accumulation of skip connections by indicators or manual design. These methods, however, are susceptible to human priors and hyper-parameters. In this work, we suggest a more subtle and direct approach that no longer explicitly searches for skip connections in the search stage, based on the paradox that skip connections were proposed to guarantee the performance of very deep networks, but the networks in the search stage of differentiable architecture search are actually very shallow. Instead, by introducing channel importance ranking and channel allocation strategy, the skip connections are implicitly searched and automatically refilled unimportant channels in the evaluation stage. Our method, dubbed Adaptive Channel Allocation (ACA) strategy, is a general-purpose approach for differentiable architecture search, which universally works in DARTS variants without introducing human priors, indicators, or hyper-parameters. Extensive experiments on various datasets and DARTS variants verify that the ACA strategy is the most effective one among existing methods in improving robustness and dealing with the collapse issue when training epochs become large.
\end{abstract}


\section{Introduction}
\label{sec:intro}

\input{01Intro}

\section{Related Work}
\label{sec:related_work}
\input{02RW}

\section{Methodology}
\label{sec:method}
\input{03Method}

\section{Experiments}
\label{sec:experiments}
\input{04Exp}

\label{sec:space}
\input{05Analysis}

\section{Conclusion}
\label{sec.Conclusion}
In this work, we proposed a novel adaptive channel allocation strategy to enhance the robustness of differentiable architecture search and addressed the architecture degradation issue of differentiable architecture search methods. Our motivation lies in the paradox that skip connections were proposed to guarantee the performance of very deep networks but networks in the search stage of differentiable architecture search are actually very shallow.
We removed the skip connection operation from the search space and then implicitly search it by adaptive channel allocation strategy based on channel importance. In this way, the adaptive channel allocation strategy can universally work in DARTS variants to achieve a stable search with high accuracy without introducing extra handcrafted prior, indicators, or hyper-parameters.

Our work conveys two important messages for future research. On the one hand, the skip connection may be unsuitable to be explicitly searched in the search phase of differentiable architecture search methods as it obviously increases the collapse risk. On the other hand, the importance of the searched operations is unequal, raising the prospect of operation-wise attention.


{\small
\bibliographystyle{IEEEtran}
\bibliography{egbib}
}

\vfill
\vfill
\vfill

\end{document}

%% file: math_commands.tex
\def\ie{\emph{i.e}.}

\def\etal{\emph{et al}.} %

\usepackage{amsmath,amsfonts,bm}

\def\eqref#1{equation~\ref{#1}}
\def\Eqref#1{Equation~\ref{#1}}

\def\1{\bm{1}}

\def\vx{{\bm{x}}}

\DeclareMathAlphabet{\mathsfit}{\encodingdefault}{\sfdefault}{m}{sl}
\SetMathAlphabet{\mathsfit}{bold}{\encodingdefault}{\sfdefault}{bx}{n}



%% file: 01Intro.tex
Neural Architecture Search (NAS) has recently attracted enormous attention since it has brought up the prospect of automating the customization of neural architecture for specific tasks. 
Compared to traditional reinforcement learning-based NAS and evolution-based NAS, Differentiable ARchiTecture Search (DARTS) reduces the search cost to the same order of magnitude as training a single neural network, enabling NAS to be applied to more expensive problems than image classification, such as semantic segmentation~\cite{liu2019auto}, disparity estimation~\cite{saikia2019autodispnet}, etc.

DARTS employs an elegant gradient-based search framework to perform differentiable joint optimization between the operation strength and the super-net weights~\cite{liu2018darts}, and thus combines the super-net search and candidate sub-network evaluation into one step. It creates a ripple effect in the NAS community and motivates a slew of new initiatives to make further improvements~\cite{cai2018proxylessnas,chang2019data,chen2019progressive,xu2019pc,DBLP:journals/tnn/ZhouXK22}. 
 
Despite its simplicity and computational efficiency, DARTS has been challenged for lacking resilience and stability. 
Many studies~\cite{zela2019understanding,liang2019darts+,chu2020fair,chu2020darts} have found that the derived target-net often collapses in the evaluation stage when \textit{skip connection} dominates the created architecture, especially after a long search stage~\cite{bi2019stabilizing}, which is well-known as the collapse issue~\cite{chu2020darts}. 
Accordingly, researchers attempt to restrict the accumulation of skip connections by adding human priors, designing indicators, or using hyper-parameters. P-DARTS~\cite{chen2019progressive} and DARTS$+$~\cite{liang2019darts+} alleviate the collapse issue by directly constraining the number of skip connections to a fixed number, which is a strong handcrafted prior. R-DARTS~\cite{zela2019understanding} indicates that the validation loss concerning the architecture parameters has higher Hessian eigenvalues when the derived architecture generalizes poorly. So R-DARTS~\cite{zela2019understanding} and SDARTS~\cite{chen2020stabilizing} attempt to enhance the robustness by regularizing for a lower Hessian eigenvalue. The main disadvantage of indicator-based approaches is that they impose strong priors by directly manipulating the derived model, which is dubious and equivalent to touching the test set~\cite{chu2020darts}. DARTS$-$~\cite{chu2020darts}, on the other hand, attempts to factor out the advantage of the skip connection with an auxiliary skip connection. However, the auxiliary skip connection introduces additional hyper-parameters. $\beta$-DARTS~\cite{ye2022beta} uses beta-decay regularization to balance the importance of different operations. 

Since the skip connection is unbeneficial for a robust and stable search, existing works aim to restrict the overaccumulation of skip connections in the search phase. However, we argue the reason might be that skip connections are unsuitable to be explicitly searched in the search stage. Skip connections were first proposed to construct a residual block with convolutions that could be used to build deeper networks with hundreds of layers without sacrificing the performance~\cite{he2016deep}. However, the super-net in differentiable architecture search is shallow, with only eight stacked cells. So it appears illogical to search the position of skip connections in the shallow super-net and then construct a deep target-net based on the searched cells. The strong evidence is that, as reported by many works~\cite{zela2019understanding,bi2019stabilizing}, DARTS can find a sound architecture by the first 50 epochs but converges to a dramatically poor architecture, with almost all operations being skip connections after searching for standard epochs (200 epochs), leading to a contradiction that skip connections are the most important operations for shallow networks. However, such architecture performs poorly in the evaluation stage, as verified in our followup experiments.

As a result,  
we conjecture if it is possible to no longer explicitly search for the skip connection in the search phase. 
Intuitively, removing the skip connection will lead to degradation in the performance of the searched architecture as the skip connection is an important candidate operation in the search space of most NAS works~\cite{cai2018proxylessnas,chang2019data,chen2019progressive,xie2018snas,xu2019pc}. So we conducted a preliminary experiment by simply removing the skip connection from the search stage of DARTS and randomly refilling them with fixed allocated channels in the evaluation stage. Surprisingly, the results exhibit excellent stability without any performance degradation.

This preliminary experiment encourages us to propose a new Adaptive Channel Allocation (ACA) strategy to implicitly search and automatically refill the skip connections to unimportant channels in the evaluation stage, where unimportant channels indicate channels that have less impact if removed from the model. Specifically, the channels are first ranked based on the channel's importance, and then the unimportant channels are refilled by skip connections. The ratio of channels of refilled skip connections is determined by the operation strength consistent with the search stage. 
Thus, the position of skip connections could be automatically determined, and a stable search could be achieved without performance degradation.

Our main contributions are as follows:
\begin{itemize}[leftmargin=20pt]
	\vspace{-1mm}
	\item We propose a novel adaptive channel allocation (ACA) strategy that first implicitly searches the skip connection to boost robust differentiable architecture search without introducing extra handcrafted prior, indicators or hyper-parameters. 
	\item 
        We design a measurement study with experiments to show that it is unnecessary to explicitly search the position of skip connections in cell-based differentiable architecture search, and the severe degradation with training epochs increasing can be solved by simply removing skip connections from the search stage and refilling them to the target-net.
	\item The proposed ACA strategy is a general-purpose robust approach that universally works in DARTS variants without extra search cost.
	\item We conduct thorough experiments across four search spaces and three datasets to demonstrate the ACA strategy is most effective among existing methods for robustifying DARTS, which can consistently improve the performance and robustness of differentiable architecture search on large search epochs.  
	
\end{itemize}

%% file: 02RW.tex
This section reviews the previous primary efforts on neural architecture search (NAS), especially DARTS-based approaches related to stability and robustness, and highlights how our work differs.

\textbf{Neural architecture search.} NAS-RL~\cite{zoph2016neural} and MetaQNN~\cite{baker2016designing}, using reinforcement learning as a search strategy to explore the search space, are considered pioneers in the field of NAS. 
Subsequently, evolution-based methods assume the possibility of applying genetic operations to force a single architecture or a family to evolve towards better performance to search for neural architectures~\cite{xie2017genetic,real2017large,liu2017hierarchical,Elsken19}. As the above methods are often very expensive, various works aim to reduce the search costs by, \emph{e.g.}, weight sharing within search models~\cite{SaxenaV16,bender_icml,Pham18,DBLP:journals/tnn/DingCLZSC22}, 
multi-fidelity optimization~\cite{baker_accelerating_2017,li-iclr17,Falkner18,Zela18}, 
and employing network morphisms~\cite{Elsken17,cai-aaai18,cai_path-level_2018,Elsken19}. 
However, their applicability often remains restricted to simple tasks or small datasets. \Revisegreen{A recent line of DARTS-based works focuses on relaxing the discrete NAS problem to a continuous one by performing differentiable joint optimization between the architecture parameters and the super-net weights and thus combines searching super-net and evaluating candidate sub-networks into one step~\cite{liu2018darts,chen2019progressive,xu2019pc,yang2021towards, gu2021dots,  garg2022learning}.} 

\textbf{Stable DARTS variants.}
Despite the simplicity of DARTS-based approaches, DARTS~\cite{liu2018darts} is known to be unstable due to the performance collapse~\cite{chang2019data,xie2018snas,zela2019understanding}. Some recent works find that the performance collapse is due to the domination of skip connection in the generated architecture~\cite{zela2019understanding,chu2020fair,chu2020darts}. Existing works have been devoted to resolving it by restricting or transferring the advantage of skip connections. Some works~\cite{zela2019understanding,chen2020stabilizing} design indicators like Hessian eigenvalues or add perturbations to regularize the indicators. However, these methods rely heavily on the accuracy of the indicators. P-DARTS~\cite{chen2019progressive} and DARTS+~\cite{liang2019darts+} employ a strong human prior to reducing the collapse issue by directly constraining the number of skip connections to a fixed number. Fair DARTS~\cite{chu2020fair} argues that the unfair advantage of skip connections lies in exclusive competition and changes the \textit{softmax activation} to \textit{sigmoid activation} to reduce competition. 
However, Fair DARTS can not limit the unfair advantage of skip connections that they often lead to the fastest way of gradient descent. SGAS~\cite{li2020sgas} employs a greedy strategy where the unfair advantage of skip connections could be prevented from taking effect. 
But greedy underestimation may prune out potential good operations too early. DARTS$-$~\cite{chu2020darts} attempts to factor out the advantage of skip connections with an auxiliary skip connection. $\beta$-DARTS, introduced by Ye et al.~\cite{ye2022beta}, employs beta-decay regularization to balance the significance of various operations. \Revisegreen{Single-DARTS~\cite{hou2021single} uses single-level optimization to update network weights and architecture parameters simultaneously, which obviously alleviates performance collapse.}

All these works attempt to restrict the skip connection in the final derived target-net by elaborate designs, while we aim to employ a direct and effective approach to stabilize DARTS by removing skip connections from the search stage and automatically refilling them in the evaluation stage.
\input{fig/darts}

%% file: fig/darts.tex
\begin{figure}[!t]
	\centering
\includegraphics[width=1\linewidth]{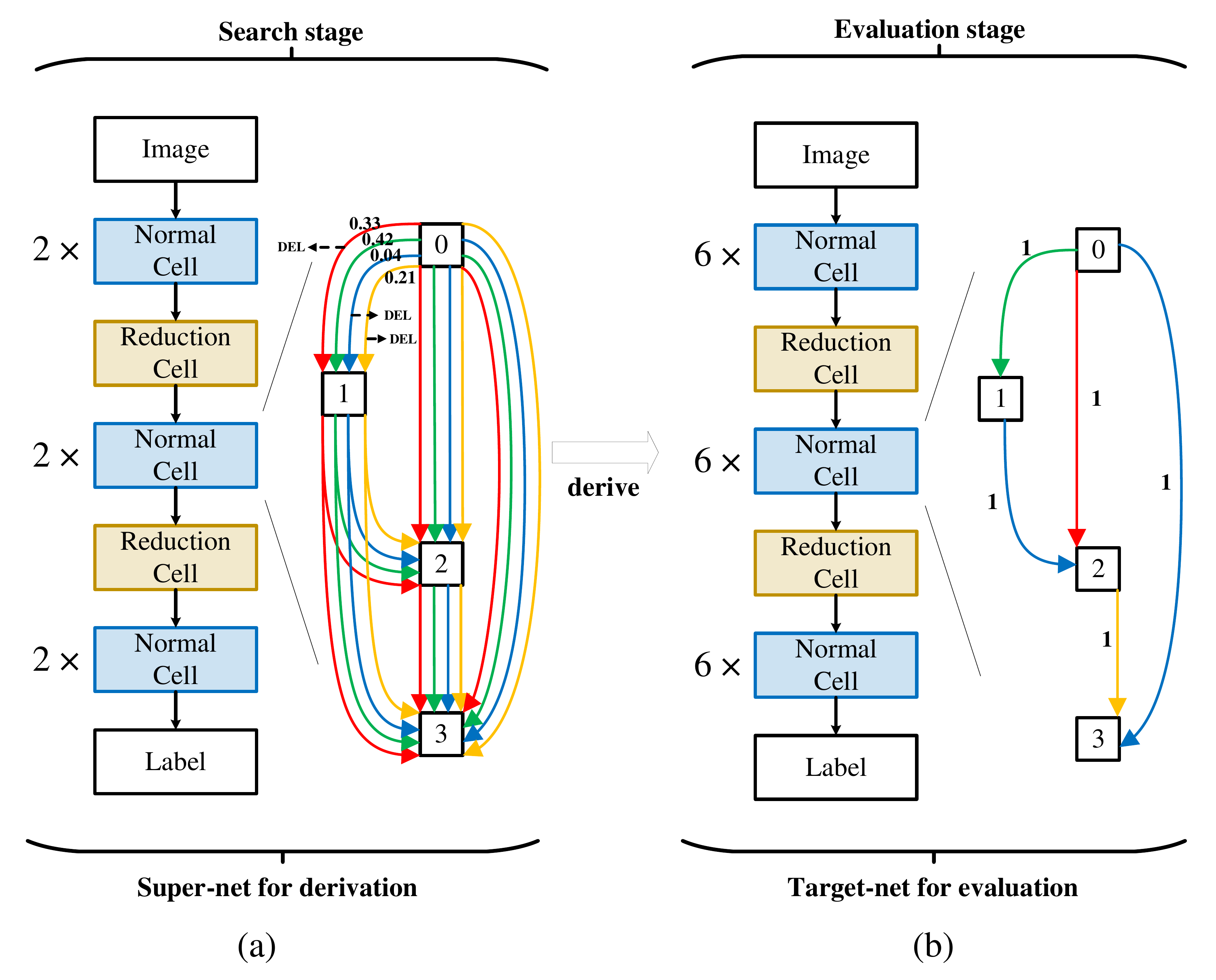}
		\caption{The overall pipeline of DARTS. (a) Super-net in the search stage. (b) Target-net in the evaluation stage.}
		\label{fig:darts}
\end{figure}

%% file: 03Method.tex
In this section, we first briefly review the differentiable architecture search widely adopted in existing works. Then we design some preliminary experiments to explore the relationship between the collapse issue and the accumulation of skip connections. Based on the results, we develop our final method of adaptive channel allocation to achieve a robust differentiable architecture search.

\subsection{Preliminaries}
DARTS-based approaches search the architectures in cell-based search space~\cite{liu2018darts,chen2019progressive,xu2019pc}, represented by a directed acyclic graph (DAG) composed of a set of nodes $N=\{\vx_1,\cdots,\vx_i,\cdots,\vx_j,\cdots,\vx_{N}\}$ and a set of edges $E= 
\{e^{(i,j)}|1 \le i<j\le N\}$. As illustrated in Fig.~\ref{fig:darts} (a), each node defines a latent representation (\emph{e.g.}, feature maps) and each edge contains $K$ operations (\emph{e.g.}, $3 \times 3$ convolution) in accordance with the candidate operations $\mathcal{O}=\{o_1, o_2,\cdots,o_k,\cdots,o_{K}\}$. 

To make the search space continuous, continuous relaxation is adopted to the categorical choice of a particular operation $o_k$ by a softmax over all possible operations in each edge $e^{(i,j)}$:
\begin{small}
\begin{align}
p^{(i,j)}_k = \frac{\exp\left(\alpha^{(i,j)}_k\right)}{\sum_k\exp\left(\alpha^{(i,j)}_k\right)}, 
	\label{eq:p}
\end{align}
\end{small}
where $\alpha^{(i,j)}_k$ is the architecture parameters related to operation $o_k$ in $e^{(i,j)}$ and $p^{(i,j)}_k\in(0,1]$ indicates the operation strength of operation $o_k$.
Then, the forward propagation in the cell of the super-net can be formulated as:
\begin{align}
\vx_j=\sum_{i<j}\sum^K_{k=1}p^{(i,j)}_k o_k\left(w^{(i,j)}_k,\vx_i\right),
\label{eq:forward}
\end{align}
where $w^{(i,j)}_k$ is the internal weights related to $o_k$ in $e^{(i,j)}$ and becomes none for non-parametric operations, such as max pooling. \par
\input{fig/epoch_skip}

After the search stage, as illustrated in Fig.~\ref{fig:darts} (b), the cell of the target-net is derived according to a handcrafted rule by regarding $p^{(i,j)}_k$ as the operation's importance, in which only the top-2 strongest input operations (from distinct nodes $\vx_i$) of each node $\vx_j$ are projected onto $1$, and others $0$. This target-net will be trained from scratch and finally get tested after retraining in the evaluation stage. 

The forward propagation of the target-net for evaluation is formulated as follows:
\begin{small}
\begin{align}
\vx_j=\sum_{(i,k)\in S_j}o_k(w^{(i,j)}_k,\vx_i),
\label{eq:eval}
\end{align}
\end{small}
\begin{small}
\begin{align}
S_j=\{(i,k)|A^{(i,j)}_k=1,\forall i<j, 1\le k\le K\},
\label{handcrafted}
\end{align}
\end{small}

\noindent where $A^{(i,j)}_k \in \{0,1\} $ is a binary variable that indicates whether operation $o_k$ in $e^{(i,j)}$ is retained. 
Comparing~\Eqref{eq:forward} and~\Eqref{eq:eval}, the handcrafted rule only retains edges with a fixed topology and discards operation strength $p^{(i,j)}_k$, which causes all retained operations in the target-net to have the same operation strength. However, the strength of different operations in the super-net is not the same.

\subsection{Motivation}
\label{sec:motivation}
Many works~\cite{zela2019understanding,liang2019darts+,chu2020fair,chu2020darts} have discovered that the derived architecture of differentiable architecture search frequently collapses after a long search stage and the primary reason is that \textit{skip connection} dominates the generated architecture. Fair DARTS~~\cite{chu2020fair} analyzes that the skip connection works well in the training phase as it can be in cooperation with convolutions. However, DARTS picks the top-performing operation (skip connection here) and discards its collaborator (convolution) before the evaluation stage, resulting in a degenerate model. DARTS$-$~\cite{chu2020darts} argues that the unfair advantage of the skip connection comes from  
its extra function in preventing gradient vanishing. Here, we believe that the performance degradation is due to the explicit searching of skip connections in the search phase.

Skip connections were first presented as a way to construct a residual block with convolutions so as to deepen the network to hundreds of layers without accuracy degradation~\cite{he2016deep}. The super-net in differentiable architecture search, however, is shallow, with just eight stacked cells. So, it appears unreasonable to explicitly search the position of skip connections in the shallow super-net and then construct a deep target-net based on the searched cells. 
To verify our claim, we first explore the performance of DARTS after removing skip connections from the search stage. We run DARTS five times on the original operation space and on the reduced space without skip connections, respectively. The target-nets are retrained for $600$ epochs.

The results are illustrated in Fig.~\ref{fig:epoch_skip}. We show by dashed lines the tendency of performance in Fig.~\ref{fig:epoch_skip} (a). The test error of DARTS increases rapidly with the increase of the search epochs. The corresponding percentage of skip connections in all the retained operations is shown in Fig.~\ref{fig:epoch_skip} (b). Clearly, the DARTS test error strongly correlates with the percentage of skip connections.  
With the increment of epochs, 
the unfair advantage of skip connections will gradually increase their architecture parameters. Because DARTS picks the top-one operation, skip connection dominates the final architecture. DARTS suffers severe degradation when 
skip connections dominate the searched architectures, which is in line with the observation of previous works~\cite{zela2019understanding,liang2019darts+,chu2020fair,chu2020darts}. 
Inspired by this observation, we conduct a preliminary experiment by simply removing skip connections from the search stage and simply randomly refilling them to the target-net with fixed allocated channels of $8$ in the evaluation stage. As shown by solid lines, DARTS without skip connection in the search stage (DARTS-S) has a consistent performance with the increment of searched epochs and exhibits smaller standard deviations than the original DARTS.

We can deduce from the comparison of DARTS and DARTS-S that DARTS shows better performance and much better stability after removing skip connections from the search stage and refilling them to the evaluation stage. Consequently, we can conclude that the existing framework is 
somewhat unreasonable to search the position of skip connections in the shallow super-net and then construct a deep target-net based on the searched cells.
In DARTS-S, however, the process of refilling skip connections in the evaluation stage is rough and manual. It motivates us to create a new strategy for implicitly searching and automatically refilling skip connections in the proper positions in the evaluation stage.


\subsection{The Proposed Adaptive Channel Allocation Strategy}
\input{fig/main_diagram}

\input{fig/ACA-DARTS}

In the search stage, the Adaptive Channel Allocation (ACA) strategy only removes the skip connection from the search space, allowing it to work in various DARTS variants universally. 
In the evaluation stage, ACA enables implicitly searching the position of skip connections by innovatively inheriting the operation strength ${p}^{(i,j)}_k$ of each retained operation $o_k$ in the search stage. 
Based on the value of  ${p}^{(i,j)}_k$, we assign different numbers of channels for each operation.

Following \Eqref{handcrafted}, we use $S =\{(i,j,k)|A^{(i,j)}_k=1,\forall i<j, 1\le k\le K\}$ to indicate all retained operations in the derived cell, where $A^{(i,j)}_k \in \{0,1\} $ is a binary variable that indicates whether operation $o_k$ in $e^{(i,j)}$ is retained. 
Then the relative channel ratio of $o^{(i,j)}_k$ could be formulated as the ratio of its operation strength to the maximum retained operation strength in the whole cell:
\begin{align}
q^{(i,j)}_k=\frac{  {p}^{(i,j)}_k}{\max\limits_{(i,j,k)\in S}\{ {p}^{(i,j)}_k\}} ,
\label{eq:ratio}
\end{align}
where the operation strength ${p}^{(i,j)}_k$ inherits from the search stage (\Eqref{eq:p}).

Let $C^{(i,j)}_k$ denotes the number of output channels of operation $o^{(i,j)}_k$, the allocated number of channels of $o^{(i,j)}_k$ is:
\begin{align}
\hat C^{(i,j)}_k=\lceil q^{(i,j)}_k \cdot C^{(i,j)}_k \rceil,
\label{eq:channel}
\end{align}
where $\lceil \cdot \rceil$ indicates the rounding symbol. 
As a result, stronger operations will be allocated more channels, and the strongest operations will be allocated all channels. The different numbers of channels of various operations achieve operation-wise attention.

However, allocating different numbers of channels to various operations in a network is difficult in practice. Moreover, according to \Eqref{eq:eval}, the output of different operations could be added together before being used as the input to the next node. So inconsistent channels could break the forward propagation. Consequently, we employ the skip connection to fill the missing channels of weak operations, preventing the cell structure from being broken. The allocated number of channels of skip connection in each $o_k$ is formulated as:
\begin{small}
\begin{align}
\overline C^{(i,j)}_k= C^{(i,j)}_k-\hat C^{(i,j)}_k.
\label{eq:skip_channel}
\end{align}
\end{small}


The refilled skip connections prevent the cell structure from being broken. On the other hand, as illustrated in Fig.~\ref{fig:channel}, the target-net automatically assigns a varied number of channels for the skip connection for different operations based on the inherited operation strength. Then, the super-net achieves an implicit search of the skip connection when it determines the operation strength of each operation. Thereafter, even if the skip connections are removed from the search space, the degradation in performance will not occur.

The most crucial factor for channel allocation is how to select channels for original operations and skip connections. We employ a strategy from network slimming~\cite{liu2017learning} by leveraging the scaling factors in batch normalization (BN) layers to determine the channel importance. Then, for each operation, the unimportant channels are replaced by the skip connection, and the critical channels for the operation remain unchanged. 
In deep learning, the BN layer plays a crucial role in normalizing internal activations using mini-batch statistics.
Let the input and output of a BN layer be $z_{in}$ and $z_{out}$, respectively. And $\mathcal{B}$ stands for the current mini-batch. The BN layer performs the following transformation:
\begin{gather}
	\hat{z} = \frac{z_{in} - \mu_\mathcal{B}}{\sqrt{\sigma_\mathcal{B}^2 + \epsilon}}; \ \
	z_{out} = \gamma \hat{z} + \beta.
 \label{eq:gamma}
\end{gather}
Here, $\mu_\mathcal{B}$ and $\sigma_\mathcal{B}$ represent the mean and standard deviation values of the input activations over the mini-batch $\mathcal{B}$. $\gamma$ and $\beta$ are trainable affine transformation parameters, allowing for the normalized activations to be linearly transformed to any desired scales. It is common practice to incorporate a BN layer after a convolutional layer with channel-wise scaling/shifting parameters. Therefore, we can directly utilize the scaling factors provided by the $\gamma$ parameters in the BN layers as channel importance. Because we drop out parameter-free operations like MaxPool, this strategy can be applied to all searched operations. The overall algorithm is provided in Algorithm 1.

Furthermore, as existing DARTS variants retain edges with a fixed topology and discard the operation strength in the target-net (see Fig.~\ref{fig:dart}), they only have limited diversity. The inherited ${p}^{(i,j)}_k$ relaxes the relative channel ratio of each operation in the target-net from $1$ to $(0,1]$, resulting in a large rise in the diversity of possible architectures in the search space. After introducing the ACA strategy, the architecture space of DARTS variants will be expanded at least $2 \times 10^{12}$ times.
\input{Table/algorithm}

\subsection{Fine-tuning}
Channel manipulation may temporarily lead to some accuracy loss. But this can be largely compensated by the following fine-tuning process~\cite{liu2017learning}. Following network slimming~\cite{liu2017learning}, we fine-tune the model for several epochs after refilling the skip connections.

\subsection{Discussion on Relationships with Prior Works}
Adaptive channel allocation, as the key strategy in this work, aims to address the performance collapse in differentiable neural architecture search. 
PC-DARTS employs a channel sampling strategy, which is comparable to our channel allocation strategy. Channel sampling is employed in the search stage, which aims to reduce computational overhead with a hyper-parameter $K$ requiring careful calibration. It works on multiple candidate operations of each edge by randomly sampling part of channels for forward propagation. In contrast, our channel allocation is a deriving strategy to address the performance collapse by automatically refilling skip connections based on channel importance and inherited operation strength.
Moreover, we introduce no handcrafted prior, indicator or hyper-parameter compared to previous works~~\cite{chen2019progressive,liang2019darts+,zela2019understanding,chen2020stabilizing,chu2020darts}, thus greatly reducing the burden of shifting to different variants and tasks.



By adopting the ACA strategy, the differentiable neural architecture search can be viewed as a combination of ResNet~\cite{he2016deep} and DenseNet~\cite{huang2017densely}. 
In the search stage, we combine features through summation before they are passed into the next node, which is in line with the strategy of ResNet. In the evaluation stage, we combine features of retained operations and skip connections by 
concatenation, 
which is similar to DenseNet. The reason for combining features through summation in the search phase is that the summation enables a better mixture of candidate operations for differentiable optimization. Instead, the summation is unsuitable for the evaluation stage because the operation strength could be balanced.

\subsection{Discussion on Inconsistency}
The inconsistency of existing DARTS variants comes from two aspects. First, each edge contains multiple operations during the search but a single operation during retraining. Second, operations have different operation strengths during the search but the same operation strength during retraining. Our method eliminates the second inconsistency by adding skip connections in the retraining based on the inherited operation strength in the search stage. The first inconsistency also exists in our method. However, it is hard to say the first inconsistency is increased in our method. Specifically, for existing DARTS variants, most skip connections exist in search but are removed during retraining. In contrast, for our method, the skip connection is removed from the search and exists in retraining.

%% file: fig/epoch_skip.tex
\begin{figure*}[!t]
\centering
\subfigure[]
{\includegraphics[width=0.41\linewidth]{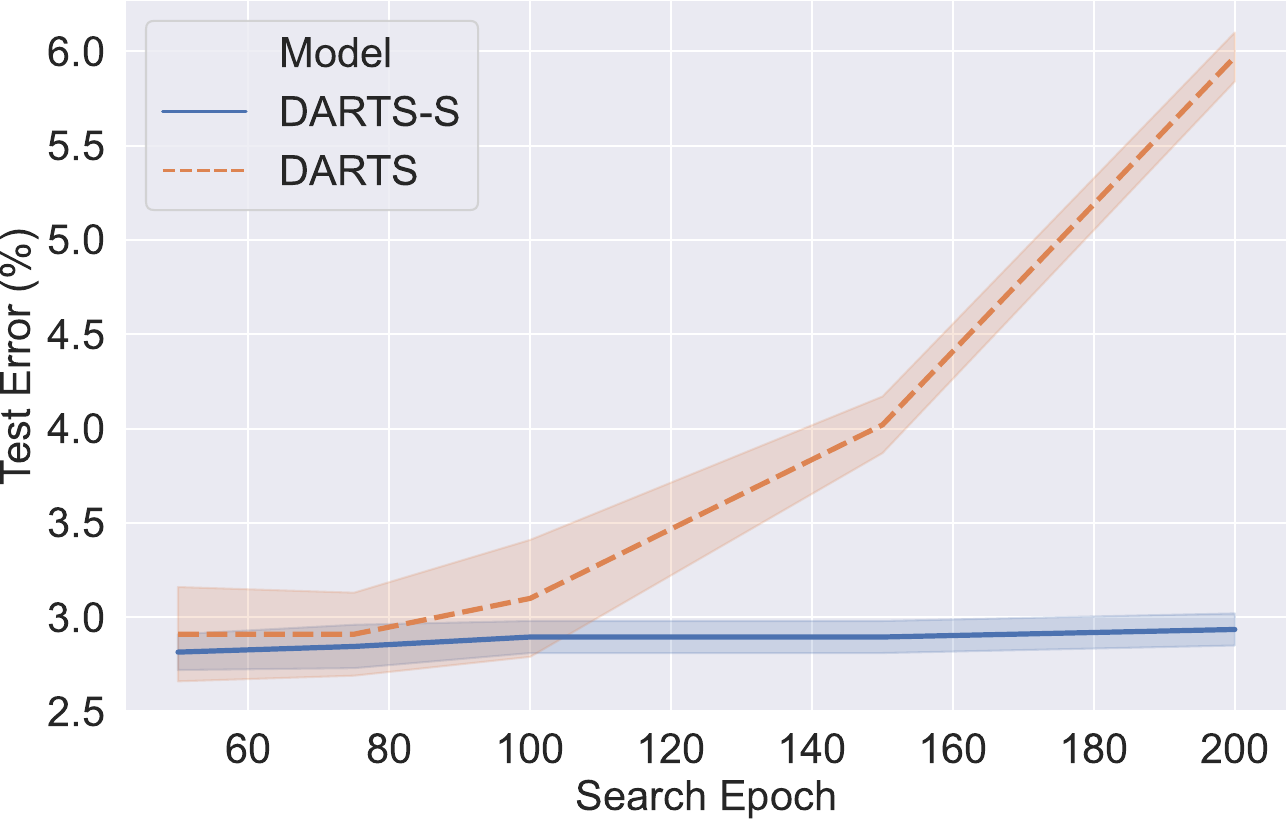}}
\hspace{10pt}
\subfigure[]
{\includegraphics[width=0.41\linewidth]{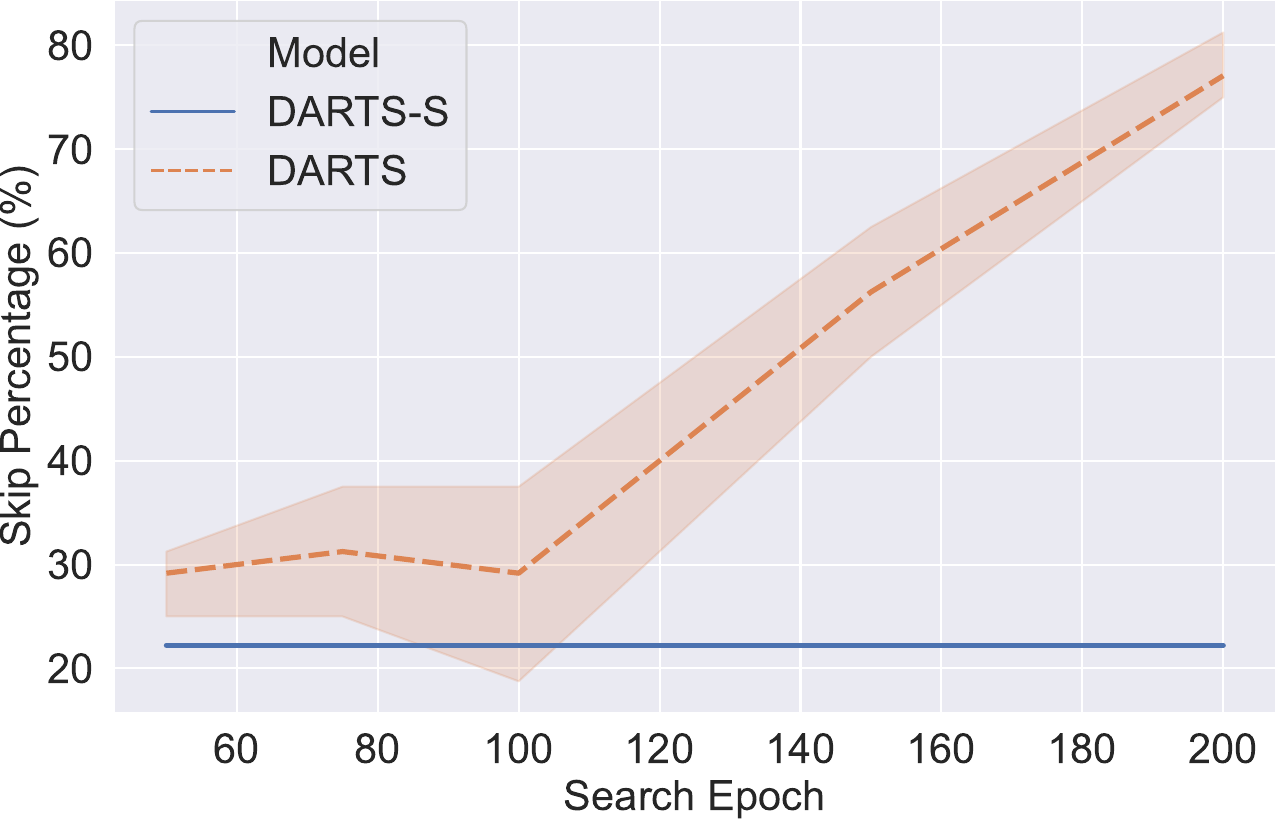}}
\caption{
The comparison of DARTS and DARTS-S (DARTS without skip connection in the search stage) on the (a) test error and (b) percentage of skip connections.
}
\label{fig:epoch_skip}
\end{figure*}

%% file: fig/main_diagram.tex
\begin{figure*}[!t]
	\centering
		\includegraphics[width=0.9\linewidth]{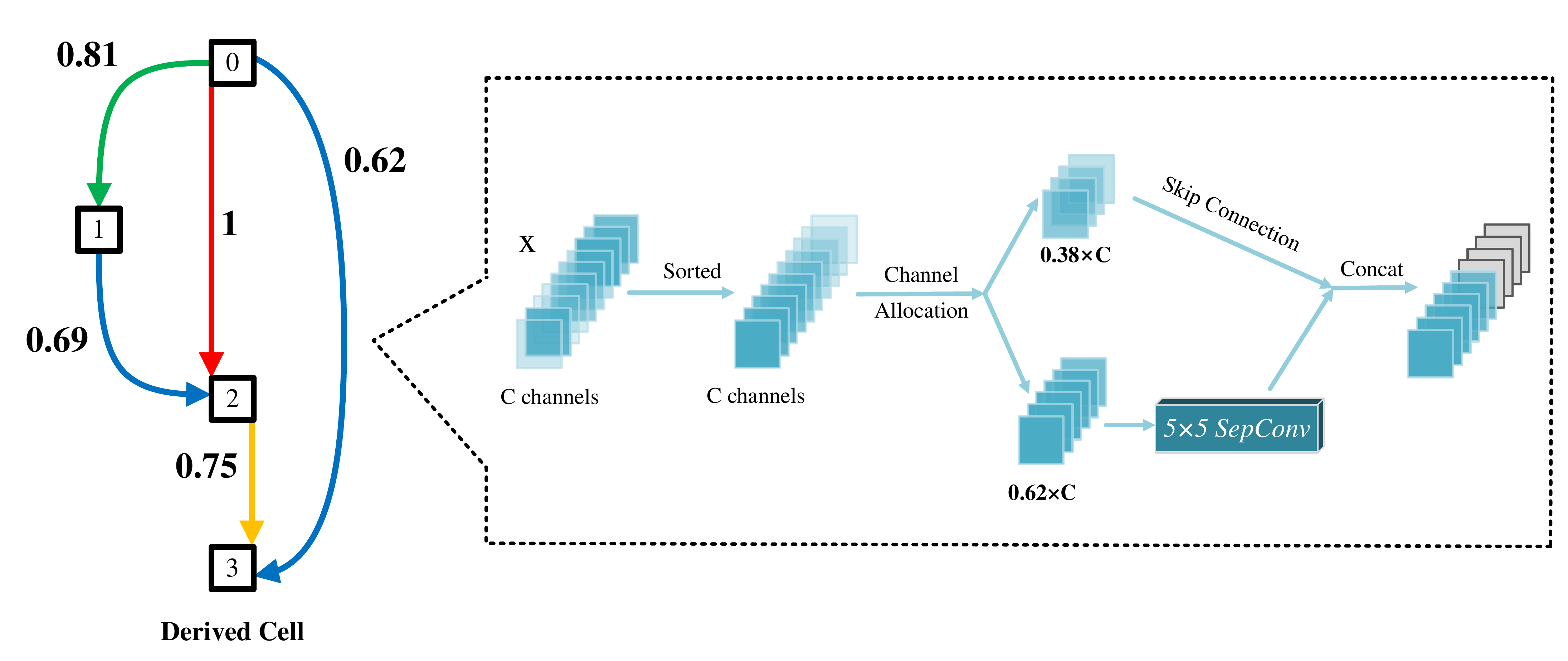}
		\caption{
  The proposed adaptive channel allocation method. In the target-net, we first employ scaling factors in batch normalization to assess the channel importance. Important channels are allocated to the original operations, and the remaining channels are assigned to skip connections. The number of allocated channels is determined by the operation strength inherited from the search stage.} 
		\label{fig:channel}
\end{figure*}

%% file: fig/ACA-DARTS.tex
\begin{figure}[!t]
	\centering
	\includegraphics[width=0.9\linewidth]{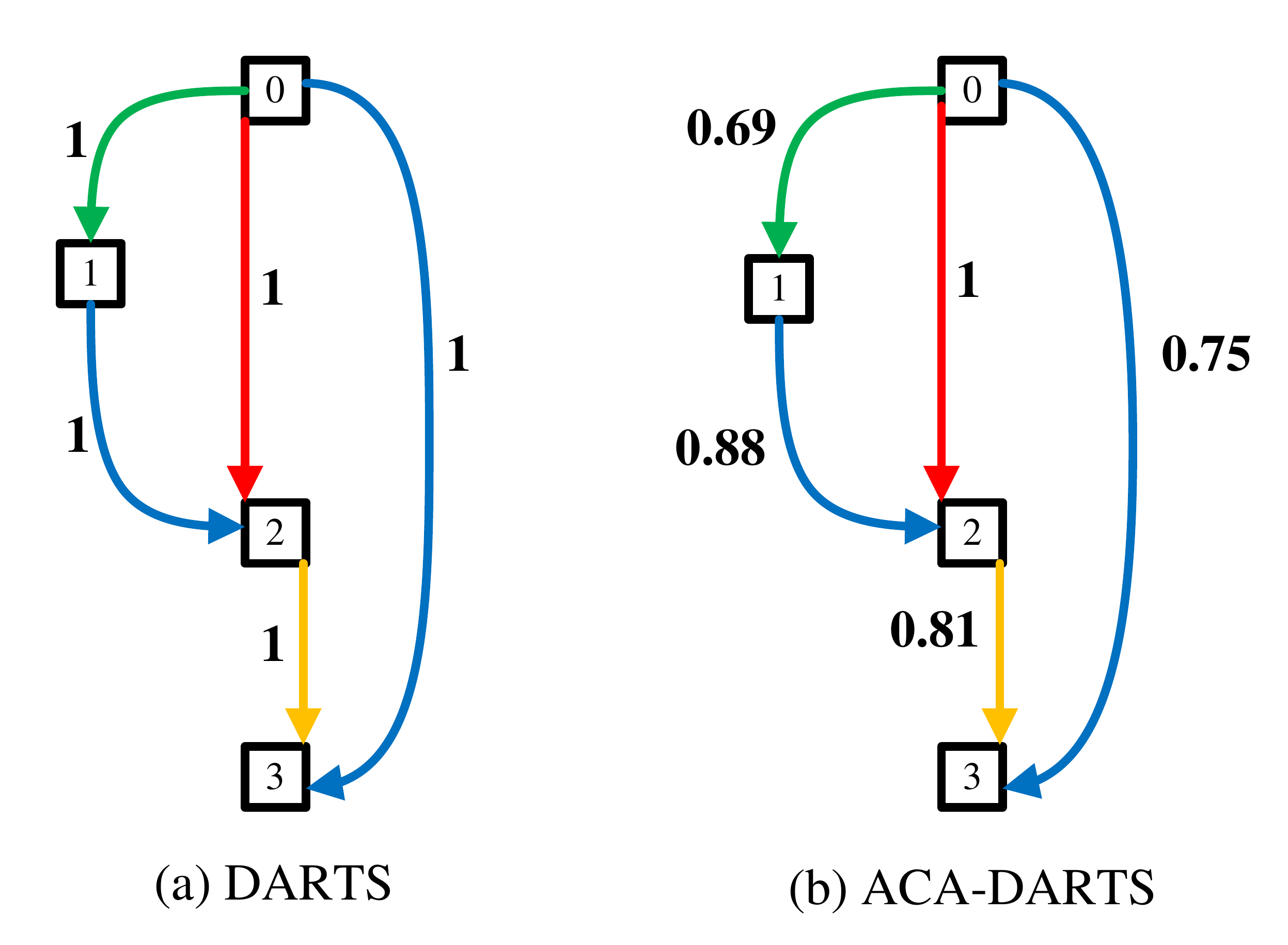}
	\caption{Schematic illustration of the cell in the target-net of (a) traditional DARTS variants and (b) the proposed ACA-based variants.}
	\label{fig:dart}
\end{figure}

%% file: Table/algorithm.tex
\begin{algorithm}[!t]
	\caption{Adaptive Channel Allocation}
	\label{alg:ACA}
	\begin{algorithmic}[1]
		\REQUIRE~~\\
		Retained operation $S =\{(i,j,k)|A^{(i,j)}_k=1,\forall i<j, 1\le k\le K\}$;\\
		Operation strength $\{p^{(i,j)}_k\}, (i,j,k)\in S$; Channel number $C$.
        \ENSURE ~~\\
            Number of channels for each retained operation $\{\hat C^{(i,j)}_k\}$;\\
            Number of channels for the skip connection $\{\overline C^{(i,j)}_k\}$.
        \STATE Determine the largest operation strength in the retained operations
 		\FOR{each operation in $S$}
		\STATE Calculate channel ratio $q^{(i,j)}_k$ based on \Eqref{eq:ratio}
		\STATE Calculate $\hat C^{(i,j)}_k$ based on \Eqref{eq:channel}
		\STATE Calculate $\overline C^{(i,j)}_k$ based on \Eqref{eq:skip_channel}
        \STATE Gain $\gamma^{(i,j)}_k$ for each channel based on \Eqref{eq:gamma}
		\ENDFOR
		\STATE Return $\{\hat C^{(i,j)}_k\}$, $\{\overline C^{(i,j)}_k\}$, $\{\gamma^{(i,j)}_k\}$.
	\end{algorithmic}
\end{algorithm}

%% file: 04Exp.tex
In this section, we first perform experiments on CIFAR-10 to investigate the combination of our approaches with DARTS and representative variants for robustifying DARTS in Section~\ref{sec:cifar10}. Then we perform experiments using a large search epoch for a thorough and comprehensive analysis of search stability (Section~\ref{sec:long}). We also show our results on ImageNet (Section~\ref{sec:imagenet}). In Section~\ref{sec:visualization}, we illustrate the visualization of our searched cells. In the end, we investigate the robustness of our methods in different search spaces, including cell space (Section~\ref{sec:cell_space}) and operation space (Section~\ref{sec:op_space}).

\input{Table/NAS-cifar10}

\subsection{Experiments on CIFAR-10 under Standard Search Space}
\label{sec:cifar10}
\textbf{Setting.} 
Following the convention~\cite{liu2018darts, xu2019pc, chen2019progressive, zela2019understanding}, we launch extensive experiments on CIFAR-10~\cite{krizhevsky2009learning} dataset.
CIFAR-10 contains $60\mathrm{K}$ images, consisting of $50\mathrm{K}$ training and $10\mathrm{K}$ testing images. These images are equally distributed over $10$ classes and each of them has a spatial resolution of $32\times32$.
Based on the observation in Section \ref{sec:motivation}, we employ adaptive channel allocation (ACA) to combine with DARTS to search convolution cells on CIFAR-10 after removing skip connections from the standard operation space. The macro architecture is constructed by stacking the cells 8 times in the search stage and 20 times in the evaluation stage. Each cell contains 7 nodes (2 input nodes, 4 intermediate nodes, and 1 output node). For a fair comparison, we follow the original setting of DARTS~\cite{liu2018darts} for detailed settings for the search and evaluation stages. Specifically, in the search stage, the network weights are optimized by momentum Stochastic Gradient Descent, with a learning rate annealed down to zero following a cosine schedule without restart, a momentum of 0.9, and a weight decay of $3\times10^{-4}$. In the evaluation stage, the models are trained from scratch for $600$ epochs and then finetuned for $10$ epochs.  

To demonstrate the universal effectiveness of the proposed ACA strategy, we also test it on popular DARTS variants aimed at improving stability or robustness, including P-DARTS~\cite{chen2019progressive} (shown as ACA-P-DARTS), PC-DARTS~\cite{xu2019pc} (shown as ACA-PC-DARTS), R-DARTS~\cite{zela2019understanding} (shown as ACA-R-DARTS), SDARTS-ADV~\cite{chen2020stabilizing} (shown as ACA-SDARTS-ADV), DARTS-~\cite{chu2020darts} (shown as ACA-DARTS-), and $\beta$-DARTS~\cite{ye2022beta} (shown as ACA-$\beta$-DARTS). We show the results of all models after searching for $50$ epochs as well as $200$ epochs to further explore the stability. All the average results are obtained on $5$ independently searched models. We can not compare the performance on architecture datasets of NAS Benchmarks~\cite{ying2019bench,DBLP:conf/iclr/ZelaSH20,DBLP:conf/iclr/Dong020} as they only contain architectures consisting of standard operations. 

\textbf{Results.} 
Table~\ref{NAS-search-CIFAR10} shows the comparison of our method with DARTS and state-of-the-art algorithms aimed at stabilizing DARTS. Compared to DARTS (1st), adaptive channel allocation (ACA-DARTS (1st)) decreases the test error from $3.00\% $ to $2.78\%$ after searching for $50$ epochs, and significantly decreases the test error from $5.41\% $ to $2.76\%$ after searching for $200$ epochs.  
When applied to other variants, the adaptive channel allocation technique achieves performance improvements for all baselines. 
After searching for 200 epochs, the improvements become significant, even if the baselines are designed for robustifying DARTS.

\subsection{Performance with Longer Epochs}
\label{sec:long}
\input{fig/large_epoch}
\textbf{Setting.} 
Experiments on CIFAR-10 show that the ACA strategy can improve the performance of DARTS and the variants aimed at strengthening stability after searching for large epochs. Here, we explore larger epochs to allow a thorough and comprehensive analysis of the search stability. Specifically, we compare ACA-DARTS (2nd) with DARTS (2nd)~\cite{liu2018darts} to demonstrate the effectiveness of our ACA strategy. Additionally, we the results of PC-DARTS~\cite{xu2019pc}, R-DARTS~\cite{zela2019understanding}, SDARTS-ADV~\cite{chen2020stabilizing}, DARTS-~\cite{chu2020darts}, and $\beta$-DARTS~\cite{ye2022beta} for comparison. P-DARTS~\cite{chen2019progressive} is not included because of its strong human prior. We run every NAS algorithm for 50, 100, 200, 300, 600, and 1000 epochs. Note that such large epochs are far from overfitting because there is still much room for improvement in the training accuracy. 

\textbf{Results} 
As illustrated in Fig.~\ref{fig:large_epoch}, DARTS (2nd)~\cite{liu2018darts} generates architectures with deteriorating performance as the search epoch becomes large, which is in line with the observation in previous works ~\cite{zela2019understanding,liang2019darts+,chu2020fair,chu2020darts}. Indicators introduced by R-DARTS~\cite{zela2019understanding}, channel sampling in PC-DARTS~\cite{xu2019pc} and auxiliary skip connection in DARTS-~\cite{chu2020darts} take some effects. However, they are only effective for the first few search epochs. $\beta$-DARTS~\cite{ye2022beta} is the most effective baseline, but it also suffers from performance degeneration after the search epochs are larger than 300. In contrast, 
ACA-DARTS (2nd) can successfully survive and overcome the instability issue even for 1000 epochs without introducing extra handcrafted prior, indicators, or hyper-parameters. Note that such strong robustness on large epochs does not occur for any other methods.
\input{fig/ACA_search_long}

\input{Table/new_space_cifar10}

The architectures derived after searching for large epochs are visualized in Fig.~\ref{fig:ACA-DARTS_long}. \Revisegreen{All architectures are constructed by stacking the cells 8 times in the search stage and 20 times in the evaluation stage. Each cell contains 7 nodes (2 input nodes, 4 intermediate nodes, and 1 output node). The first and second nodes of cell $k$ are set equal to the outputs of cell $k-2$ and cell $k-1$, respectively. Cells located at the $1/3$ and $2/3$ of the total depth of the network are reduction cells, in which all the operations adjacent to the input nodes are of stride two. And other cells are normal cells.} 
The number behind each operation in a cell represents its channel ratio. The channel ratio calculates the allocated number of channels, and the remaining channels are allocated to skip connections. It is worth noting that all the architectures are reasonable, without the domination of skip connections in the final architectures.

\subsection{\Revise{Experiments on CIFAR-10 under Complex Search Space}}

\Revise{\textbf{Search space.} Traditionally, DARTS and its variations have relied on numerous manually crafted rules to determine the ultimate architecture. These rules include constraints such as each edge preserving only one operator, each inner node retaining two of its predecessors, and the architecture being shared among cells of the same type (normal and reduction). While these constraints contribute to the stability of the search process, they also constrain the flexibility of architecture exploration. For instance, the requirement for low-level and high-level layers to have identical topological complexity may not always yield optimal solutions. 
A recent work~\cite{bi2019stabilizing} highlights the importance of evaluating NAS approaches within a more intricate search space, with fewer heuristic rules. Motivated by this, GOLD-NAS~\cite{bi2020gold} relaxes the heuristic constraints, providing higher flexibility to the final architecture. Specifically, it allows each edge to retain an arbitrary number of operators (directly summed into the output), each inner node to preserve an arbitrary number of predecessors and render all cell architectures independent. To accommodate limited GPU memory, it adopts the strategy of preserving two operators, \emph{SkipConnect} and $3\times3$ \emph{SepConv}, in each edge.}

\Revise{It is crucial to recognize that reducing the number of candidate operators does not simplify the search task. Even with only two candidates per edge, the expanded space remains vast, estimated at $3.1\times10^{117}$ architectures, far surpassing the capacity of the original space with $1.1\times10^{18}$ architectures.  Navigating this enlarged space without heuristic rules demands more sophisticated search methods. }

\Revise{\textbf{Setting.} Following GOLD-NAS, we independently search 14 cells in the search stage and use the searched 14 cells for the re-train procedure. The initial channels before the first cell are set to 36.  
During the search, architectural parameters are initialized to zeros, with a batch size of 96. 
We integrate  adaptive channel allocation (ACA) with various NAS methods, including DARTS~\cite{liu2018darts}, PC-DARTS~\cite{xu2019pc}, DARTS-~\cite{chu2020darts}, SDARTS-ADV~\cite{chen2020stabilizing}, GOLD-NAS~\cite{bi2020gold}, $\beta$-DARTS~\cite{ye2022beta}, and Single-DARTS~\cite{hou2021single}, to search cells on CIFAR-10 after excluding skip connections from the complex operation space. 
P-DARTS~\cite{chen2019progressive} and R-DARTS~\cite{zela2019understanding}, which under-performed in the original search space, are excluded due to the time limitations. Notably, except for GOLD-NAS, all methods include the \emph{Zero} operation in their search space, 
where the \emph{Zero} operation simply replaces every value in the input feature map with zeros, \ie, no connection. 
}

\Revise{\textbf{Results.}
Table~\ref{new-space-CIFAR10} presents a comparison of our method with DARTS and state-of-the-art methods aiming at stabilizing DARTS in the complex search space. Several conclusions can be drawn from these results. First, when searching for $50$ epochs, our ACA strategy can boost the performance of most algorithms. Second, when searching for $200$ epochs, our ACA strategy can significantly improve the performance of all methods. Third, all existing methods perform worse after searching for longer epochs due to the existence of \emph{SkipConnect}. Fourth, after combining with the ACA strategy, most methods perform better after searching for longer epochs, indicating that a longer search is necessary when the performance collapse issue is addressed. }

\input{Table/NAS-imagenet}

\subsection{Experiments on ImageNet under Standard Search Space}
\label{sec:imagenet}
\textbf{Setting.} 
The ImageNet~\cite{deng2009imagenet} dataset contains $1.3\mathrm{M}$ training images and $50\mathrm{K}$ validation images, all of which are high-resolution and roughly equally distributed over $1\rm{,}000$ object categories.
We test the performance of the adaptive channel allocation strategy on ImageNet. We follow the original setting of DARTS and DARTS variants for a fair comparison.
Here, the macro architecture is obtained by stacking the cells 8 times in the search stage and 14 times in the evaluation stage. We show the results of all models after searching for 50 epochs as well
as 200 epochs to further explore the stability. 
We train the network for 250 epochs by an SGD optimizer with an annealing learning rate initialized as 0.5, a weight decay of $3\times 10^{-5}$, and a momentum of 0.9. 
Similar to previous works~\cite{xu2019pc,chen2019progressive}, 
we also employ auxiliary loss of tower and label smoothing to enhance the training. 
We finetune the models with a learning rate of $10^{-3}$ for only $5$ epochs. 
Following the conventions~\cite{zoph2018learning,liu2018darts,xu2019pc}, we apply the \textit{mobile setting}, where the number of multi-add operations does not exceed $600\mathrm{M}$ and the input image size is fixed to $224\times224$.

\textbf{Results.} 
As illustrated in Table~\ref{tab:imagenet}, ACA-DARTS outperforms DARTS by a large margin. 
Furthermore, the adaptive channel allocation strategy achieves improved performance when applied to the popular DARTS variants, including P-DARTS~\cite{chen2019progressive}, PC-DARTS~\cite{xu2019pc}, R-DARTS~\cite{zela2019understanding}, SDARTS-ADV~\cite{chen2020stabilizing}, and $\beta$-DARTS~\cite{ye2022beta}. The results demonstrate the effectiveness of our method on large-scale tasks. Our best run achieves a top-1 test error of $23.9\%$, which is competitive amongst popular NAS methods. In addition, after searching for $200$ epochs, the ACA strategy can improve the performance of all baselines significantly.

\input{fig/searched_cells}
\input{fig/searched_architecture}

\subsection{Visualization on the Searched Cells}
\label{sec:visualization}
On the original search space, we visualize the searched normal cells and reduction cells of ACA-DARTS (2nd) on CIFAR-10 and ImageNet in Fig.~\ref{fig:searched_cells_DART_imagenet}. The main difference between our searched cells and those of other DARTS-based approaches is the channel ratio for each operation. The channel ratio for other approaches is fixed as one, even if the retained operations have unequal contributions. In contrast, our channel ratio is calculated by inherited operation strength ${p}^{(i,j)}_k$. Then the stronger operation will be allocated more channels, and the remaining unimportant channels are allocated to skip connections, achieving operation-wised attention and implicit skip connection search. Furthermore, the inherited ${p}^{(i,j)}_k$ increases the diversity significantly.

\Revise{On the complex search space, we visualize the searched architectures of ACA-DARTS (2nd) on CIFAR-10 in Fig.~\ref{fig:searched_architecture}. From the two architectures, we observe some interesting features. First, for the $k$-th cell, the input connections from the $(k-2)$-th cell are more than those from the $(k-1)$-th cell. Second, connections are denser at the deeper parts in both architectures. Third, the connections between cells are denser than those within cells. Fourth, channel ratios are smaller in the deeper parts of both architectures, indicating \emph{SkipConnects} are more important in deeper layers.}

\input{fig/n}

%% file: Table/NAS-cifar10.tex
\begin{table*}[t]
\centering
\caption{Comparison with state-of-the-art approaches related to robustifying DARTS on CIFAR-10.   
Each result is averaged from 5 independently searched models to represent the robustness. 
1st means using the first-order approximation. 2nd means using the second-order approximation.}
\label{NAS-search-CIFAR10}
\begin{threeparttable}[b]
\small
\resizebox{0.90\textwidth}{!}{
\begin{tabular}{@{}lccccc@{}} 	
\toprule		
\multirow{2}{*}{\textbf{Architecture}}   &  \multicolumn{2}{c}{\textbf{Test Error (\%)}}  & \textbf{Params$^*$}   &   \textbf{Search Cost}$^\dagger$& \textbf{Robustifing}  \\
& \textbf{50 epochs}&\textbf{200 epochs} & {\textbf{(M)}}  & \textbf{GPU-days}& \textbf{Method}   \\
\midrule
{DARTS} (1st)\cite{liu2018darts}   &  3.00$\pm$0.14  & 5.41$\pm$0.30  &  3.3$\pm$0.4  & 0.4  &- \\
ACA-DARTS (1st) &  \textbf{2.78$\pm$0.10} &\textbf{2.76$\pm$0.13}  & 3.5$\pm$0.2    & 0.4 & ACA\\
\midrule
{DARTS} (2nd)\cite{liu2018darts}   &  2.76$\pm$0.09  & 4.12$\pm$0.21  &  3.4$\pm$0.3  & 1.0  &- \\
ACA-DARTS (2nd) &  \textbf{2.68$\pm$0.09} &\textbf{2.69$\pm$0.11}  & 3.6$\pm$0.2    & 1.0 & ACA\\
\midrule
{PC-DARTS}\cite{xu2019pc}    &  2.57$\pm$0.07 & 3.85$\pm$0.19 &  3.6$\pm$0.3  &  0.1  & channel sampling\\ 
{ACA-PC-DARTS} &  \textbf{2.54$\pm$0.09} & \textbf{2.74$\pm$0.11} &  3.6$\pm$0.2    &  0.1  &ACA+channel sampling\\
\midrule
{P-DARTS}$^\ddagger$\cite{chen2019progressive}   &  2.81$\pm$0.14 &2.93$\pm$0.15  &  3.4$\pm$0.3  &  0.3  & human prior\\ 
{ACA-P-DARTS}    &  \textbf{2.73$\pm$0.12} & \textbf{2.76$\pm$0.11}&  3.5$\pm$0.2  &  0.3  &ACA+human prior\\
\midrule
{R-DARTS}\cite{zela2019understanding}   
&2.95$\pm$0.21 & 4.34$\pm$0.21& 3.4$\pm$0.3 & 0.4&indicators\\  
{ACA-R-DARTS}   
&\textbf{2.81$\pm$0.13} & \textbf{2.82$\pm$0.09}& 3.6$\pm$0.2 & 0.4 &ACA+indicators\\
\midrule
{DARTS-}\cite{chu2020darts}   
&  2.59$\pm$0.08 & 3.45$\pm$0.15 &  3.5$\pm$0.2  & 0.4  &hyper-parameters\\
{ACA-DARTS-}   
&  \textbf{2.59$\pm$0.09} &\textbf{2.72$\pm$0.07} &  3.5$\pm$0.1   & 0.4 &ACA+hyper-parameters \\
\midrule
{SDARTS}-ADV\cite{chen2020stabilizing}  
& 2.61$\pm$0.02 &3.14$\pm$0.14 & 3.3$\pm$0.1& 1.3&indicators\\
{ACA-SDARTS}-ADV 
&  \textbf{2.58$\pm$0.08} &\textbf{2.73$\pm$0.08}  & 3.4$\pm$0.2 & 1.2 &ACA+indicators\\
\midrule
{$\beta$-DARTS}\cite{ye2022beta}  
& 2.53$\pm$0.08 &2.79$\pm$0.11 & 3.8$\pm$0.2& 0.4&beta decay \\
{ACA-$\beta$-DARTS} 
&  \textbf{2.52$\pm$0.10} &\textbf{2.56$\pm$0.09}  & 3.8$\pm$0.2 & 0.4 &ACA+beta decay\\
\bottomrule
\end{tabular}}
\begin{tablenotes}
\footnotesize
\item[$*$] Parameters of the models after searching for 50 epochs. Parameters under 200 epochs are similar to 50 epochs.
\item[$\dagger$] Search cost for 50 epochs. The search cost of 200 epochs takes about 4 times that of 50 epochs.
\item[$\ddagger$] 5 independent searches using their released code.
\end{tablenotes}
\end{threeparttable}
\end{table*}

%% file: fig/large_epoch.tex
\begin{figure}[!t]
	\centering
		\includegraphics[width=0.95\linewidth]{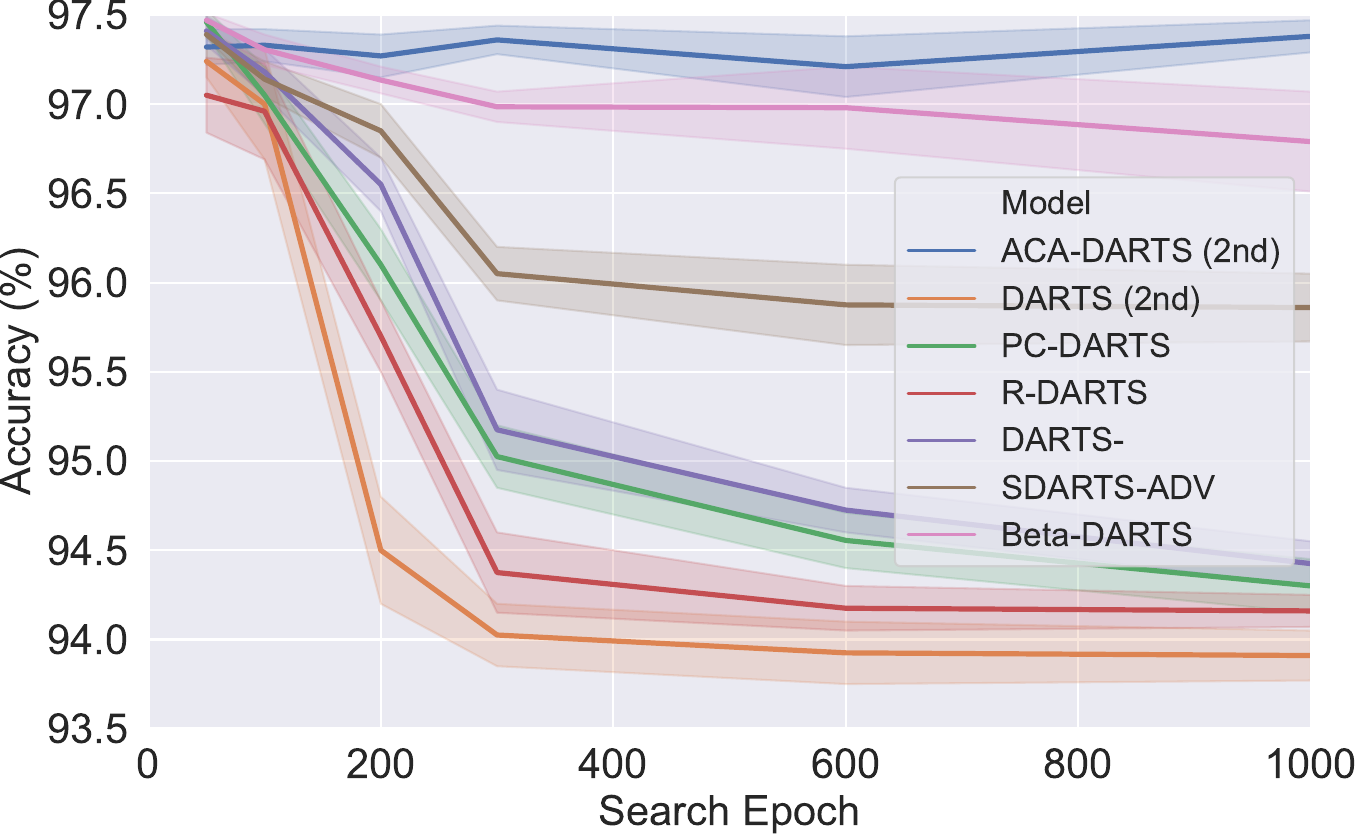}
		\caption{Test accuracy on CIFAR-10 after searching for large epochs (best view in color).}
		\label{fig:large_epoch}
\end{figure}

%% file: fig/ACA_search_long.tex
\begin{figure*}[!t]
\centering
\subfigure[Normal]{\includegraphics[width=0.18\linewidth]{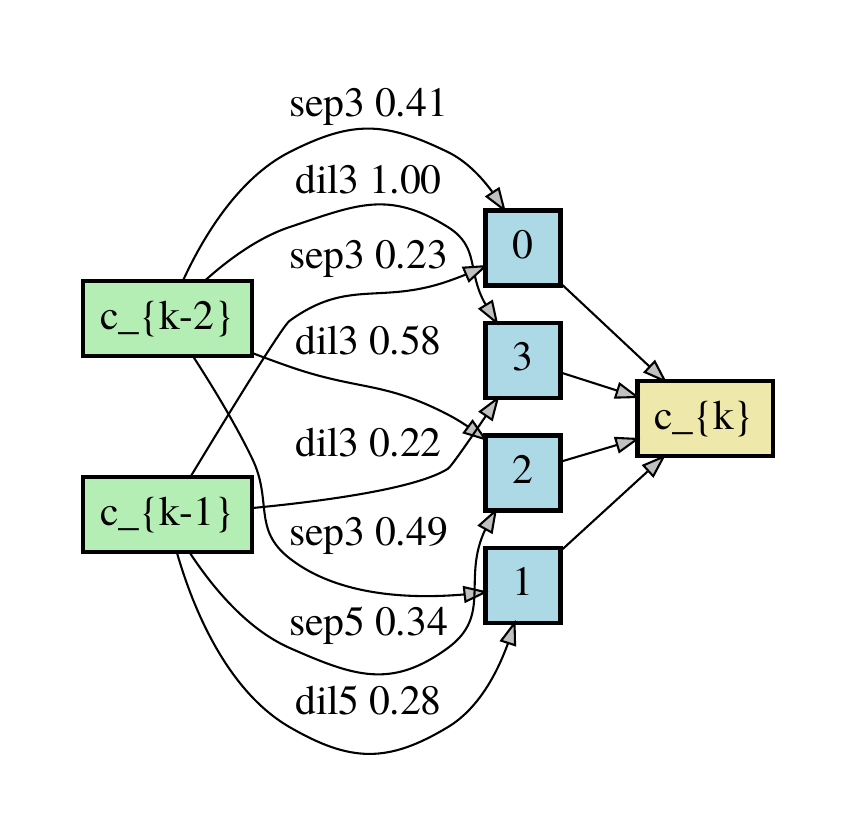}}
\hspace{0.1em}
\subfigure[Reduction]{\includegraphics[width=0.27\linewidth]{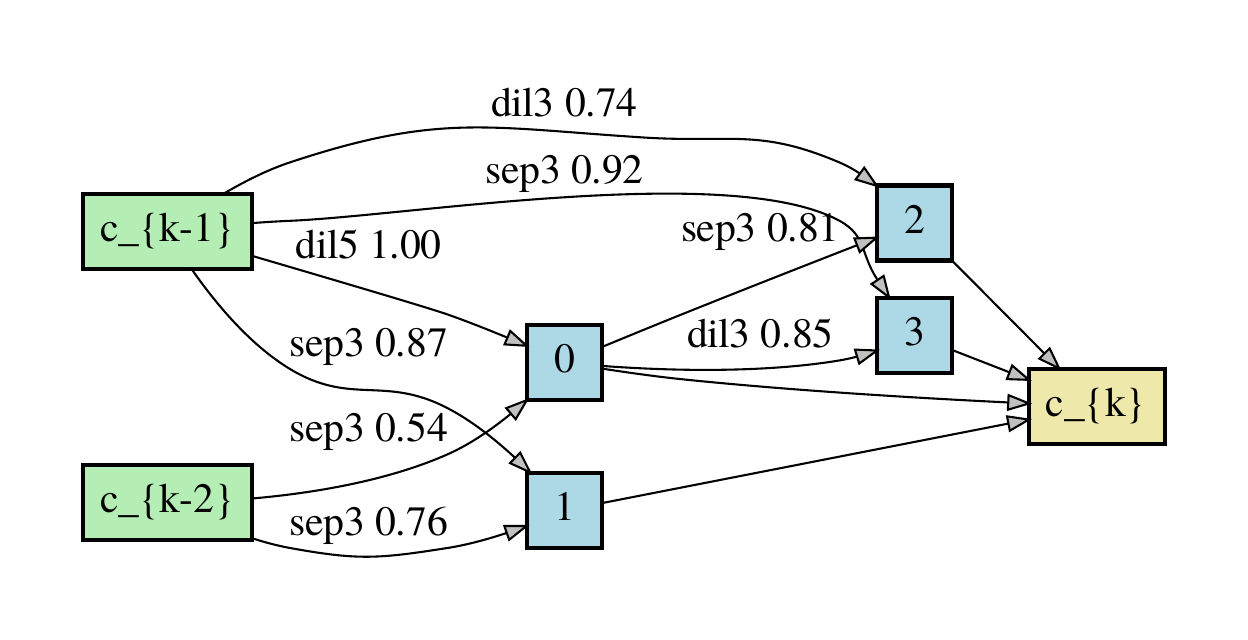}}
\vspace{-1ex} 
\subfigure[Normal]{\includegraphics[width=0.18\linewidth]{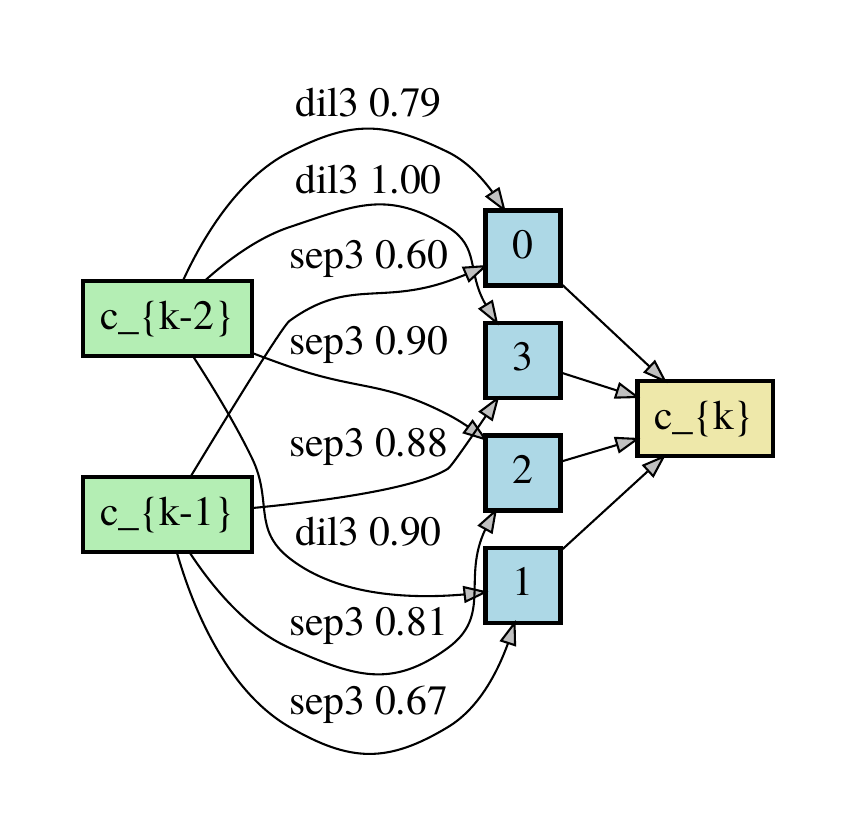}}
\hspace{0.1em}
\subfigure[Reduction]{\includegraphics[width=0.27\linewidth]{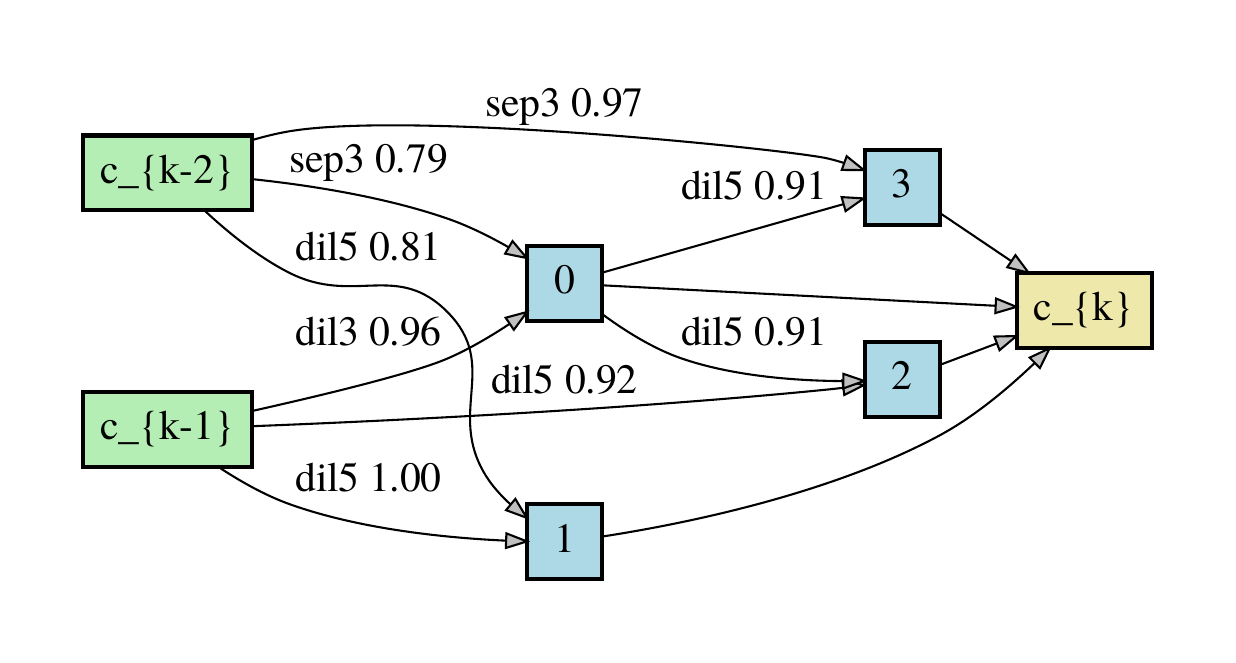}}
 \\ 
\subfigure[Normal]{\includegraphics[width=0.18\linewidth]{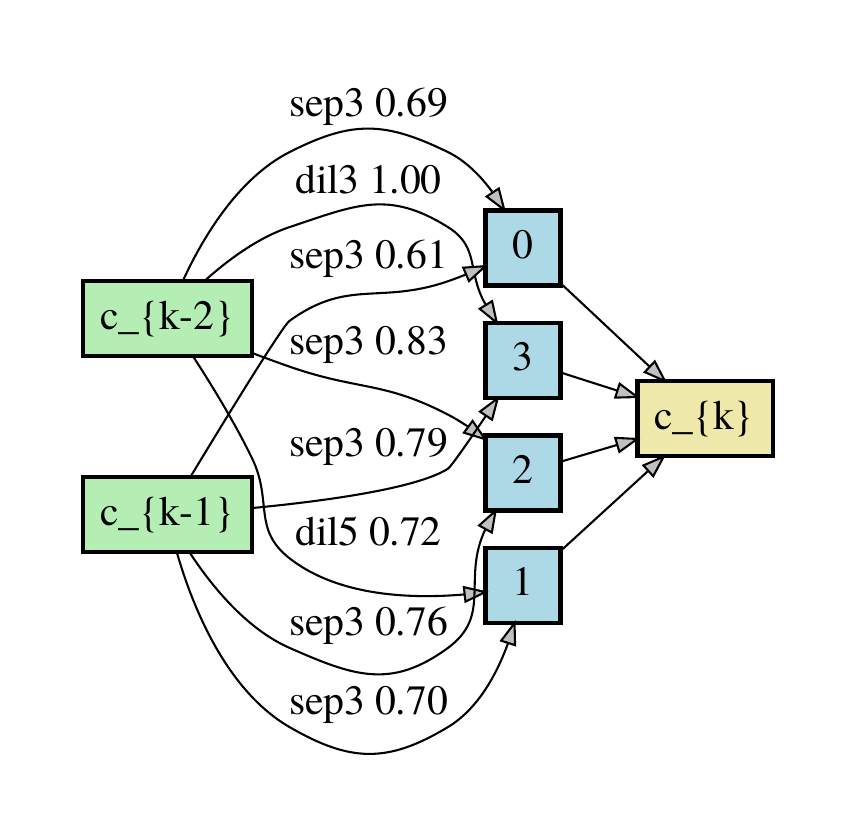}}
\hspace{0.1em}
\subfigure[Reduction]{\includegraphics[width=0.27\linewidth]{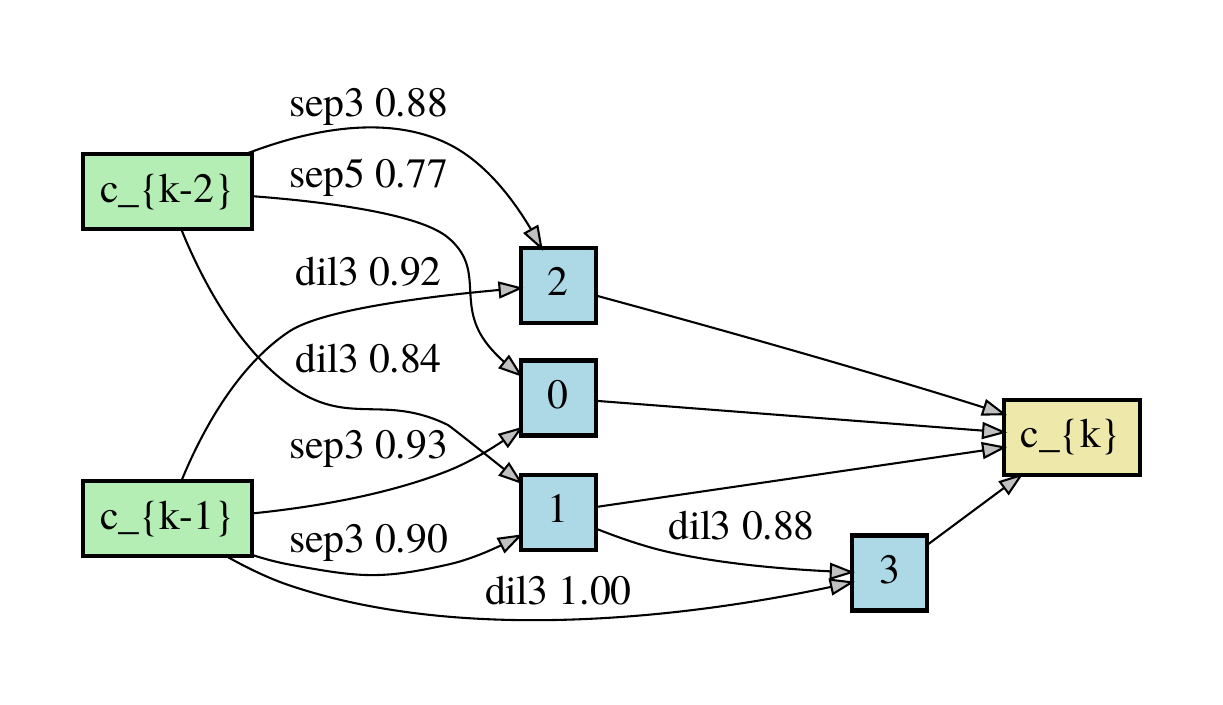}}
\vspace{-1ex} 
\subfigure[Normal]{\includegraphics[width=0.18\linewidth]{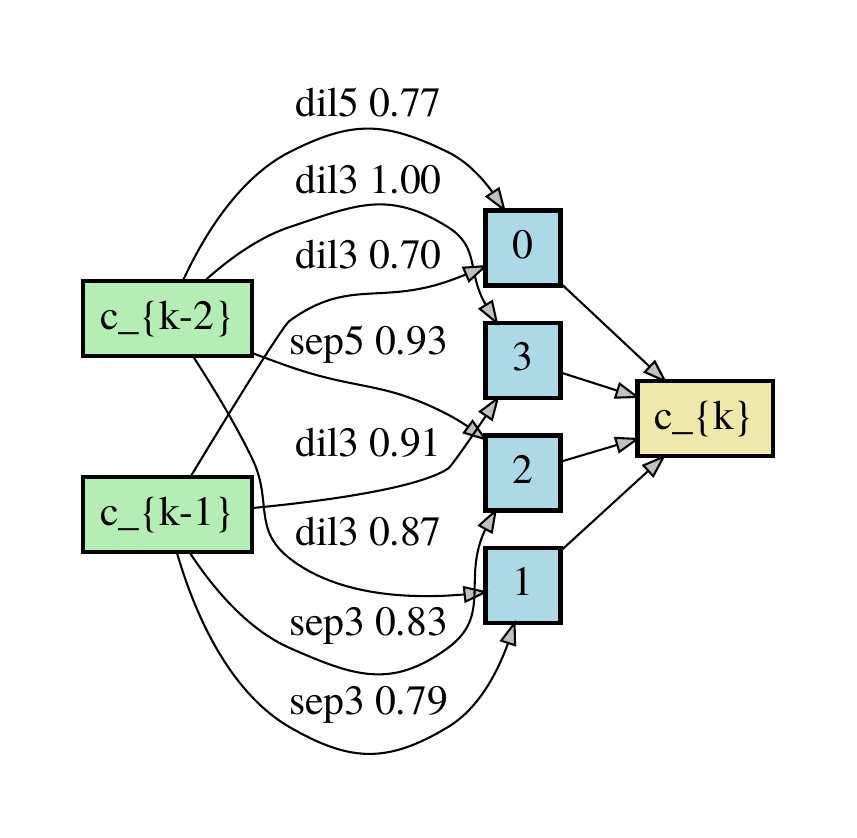}}
\hspace{0.1em}
\subfigure[Reduction]{\includegraphics[width=0.27\linewidth]{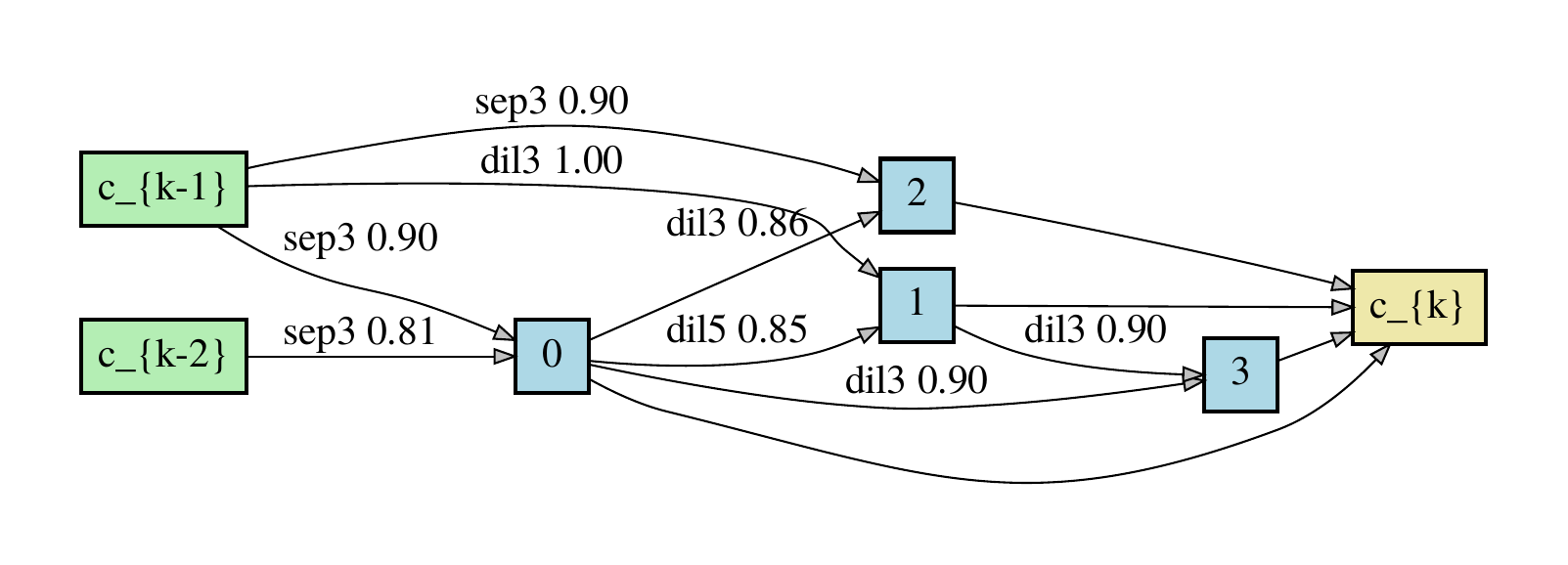}}
\caption{Searched normal cells and reduction cells by ACA-DARTS on CIFAR-10 dataset with large search epochs.
(a) Normal cell of 200 search epochs.
(b) Reduction cell of 200 search epochs. 
(c) Normal cell of 300 search epochs. 
(d) Reduction cell of 300 search epochs. 
(e) Normal cell of 600 search epochs.
(f) Reduction cell of 600 search epochs. 
(g) Normal cell of 1000 search epochs. 
(h) Reduction cell of 1000 search epochs. The number behind each operation represents its channel ratio.}
 
\label{fig:ACA-DARTS_long}
\end{figure*}

%% file: Table/new_space_cifar10.tex
\begin{table*}[!t]
\centering
\caption{\Revise{Comparison with state-of-the-art approaches related to robustifying DARTS on CIFAR-10 under the complex search space. Each result is averaged from 5 independently searched models to represent the robustness. 1st means using the first-order approximation. 2nd means using the second-order approximation.}}
\label{new-space-CIFAR10}
\begin{threeparttable}[b]
\small
\resizebox{0.90\textwidth}{!}{
\begin{tabular}{@{}lccccc@{}} 	
\toprule		
\multirow{2}{*}{\textbf{Architecture}}   &  \multicolumn{2}{c}{\textbf{Test Error (\%)}}  & \textbf{Params}   &   \textbf{Search Cost}& \textbf{Robustifing}  \\
& \textbf{50 epochs}&\textbf{200 epochs} & {\textbf{(M)}}  & \textbf{GPU-days}& \textbf{Method}   \\
\midrule
{DARTS} (1st)\cite{liu2018darts}   &  3.11$\pm$0.17  & 4.95$\pm$0.31  &  3.6$\pm$0.3  & 0.6  &- \\
ACA-DARTS (1st) &  \textbf{2.68$\pm$0.15} &\textbf{2.59$\pm$0.17}  & 3.5$\pm$0.3    & 0.4 & ACA\\
\midrule
{DARTS} (2nd)\cite{liu2018darts}   &  2.84$\pm$0.13  & 4.35$\pm$0.19  &  3.5$\pm$0.1  & 1.2  &- \\
ACA-DARTS (2nd) &  \textbf{2.58$\pm$0.12} &\textbf{2.52$\pm$0.13}  & 3.5$\pm$0.2    & 0.8 & ACA\\
\midrule
{PC-DARTS}\cite{xu2019pc}    &  2.82$\pm$0.19 & 3.49$\pm$0.17 &  3.5$\pm$0.2  &  0.2  & channel sampling\\ 
{ACA-PC-DARTS} &  \textbf{2.60$\pm$0.23} & \textbf{2.56$\pm$0.14} &  3.4$\pm$0.2    &  0.1  &ACA+channel sampling\\
\midrule
{DARTS-}\cite{chu2020darts}   
&  2.67$\pm$0.15 & 3.07$\pm$0.12 &  3.6$\pm$0.1  & 0.6  &hyper-parameters\\
{ACA-DARTS-}   
&  \textbf{2.61$\pm$0.13} &\textbf{2.57$\pm$0.08} &  3.6$\pm$0.2   & 0.4 &ACA+hyper-parameters \\
\midrule
{SDARTS}-ADV\cite{chen2020stabilizing}  
& 2.67$\pm$0.07 &3.28$\pm$0.16 & 3.5$\pm$0.2& 1.6&indicators\\
{ACA-SDARTS}-ADV 
&  \textbf{2.59$\pm$0.09} &\textbf{2.55$\pm$0.06}  & 3.6$\pm$0.3 & 1.2 &ACA+indicators\\
\midrule
{GOLD-NAS}\cite{bi2020gold}  
& 2.53$\pm$0.08 &2.60$\pm$0.17 & 3.7$\pm$0.3& 1.1&hyper-parameters \\
{ACA-GOLD-NAS} 
&  \textbf{2.53$\pm$0.11} &\textbf{2.53$\pm$0.09}  & 3.7$\pm$0.3 & 1.1 &ACA+hyper-parameters\\
\midrule
{$\beta$-DARTS}\cite{ye2022beta}  
& \textbf{2.57$\pm$0.11} &2.79$\pm$0.13 & 3.8$\pm$0.3& 0.6&beta decay \\
{ACA-$\beta$-DARTS} 
&  2.58$\pm$0.13 &\textbf{2.49$\pm$0.06}  & 3.8$\pm$0.2 & 0.4 &ACA+beta decay\\
\midrule
{Single-DARTS}\cite{hou2021single}  
& 2.58$\pm$0.10 &2.73$\pm$0.09 & 3.4$\pm$0.2& 0.6&single-level\\
{ACA-Single-DARTS} 
&  \textbf{2.54$\pm$0.14} &\textbf{2.52$\pm$0.08}  & 3.5$\pm$0.2 & 0.4 &ACA+single-level\\
\bottomrule
\end{tabular}}
\end{threeparttable}
\end{table*}

%% file: Table/NAS-imagenet.tex
\begin{table}[!t]
    \centering
    \caption{Comparison with state-of-the-art image classifiers on ImageNet in the mobile setting.}
    \begin{threeparttable}[b]
    \small
    \resizebox{0.45\textwidth}{!}{
    
    \begin{tabular}{lcc}
    
    \hline
    \multirow{2}*{\textbf{Architecture}} & \multicolumn{2}{c}{\textbf{Test Error (\%)}} \\ \cline{2-3}
    & 50 epochs & 200 epochs \\ \hline
    DARTS (2nd)\cite{liu2018darts} & 26.7 & 47.7 \\
        ACA-DARTS (2nd) & 25.0 & 24.8  \\ 
        \hline
    P-DARTS\cite{chen2019progressive} & 24.4 & 26.4 \\ 
        ACA-P-DARTS & 24.3 & 24.2  \\
        \hline
    PC-DARTS\cite{xu2019pc}  & 25.1 & 33.8 \\
        ACA-PC-DARTS & 24.5 &  24.3  \\
        \hline
    PC-DARTS$^\dagger$\cite{xu2019pc}  & 24.2 & 36.1 \\
        ACA-PC-DARTS$^\dagger$ & 24.1 & 24.3   \\
        \hline
    R-DARTS\cite{zela2019understanding} & 24.8 & 39.2 \\ 
        ACA-R-DARTS & 24.5 &  24.9   \\
        \hline
    SDARTS-ADV\cite{chen2020stabilizing} & 25.2 & 31.5 \\
        ACA-SDARTS-ADV & 24.8 & 24.6 \\
        \hline
    $\beta$-DARTS & 23.9 & 24.8 \\
        ACA-$\beta$-DARTS & 23.9 &   24.0     \\
    \hline
    \end{tabular}}

\begin{tablenotes}
\footnotesize
\item[$\dagger$] This architecture was searched on ImageNet directly. 
\end{tablenotes}
    \end{threeparttable}
    \label{tab:imagenet}
\end{table}

%% file: fig/searched_cells.tex
\begin{figure*}[!t]
\centering
\hspace{5em}
\subfigure[The normal cell on CIFAR-10]
{\label{aca_cifar_ncells}\includegraphics[width=0.27\linewidth]{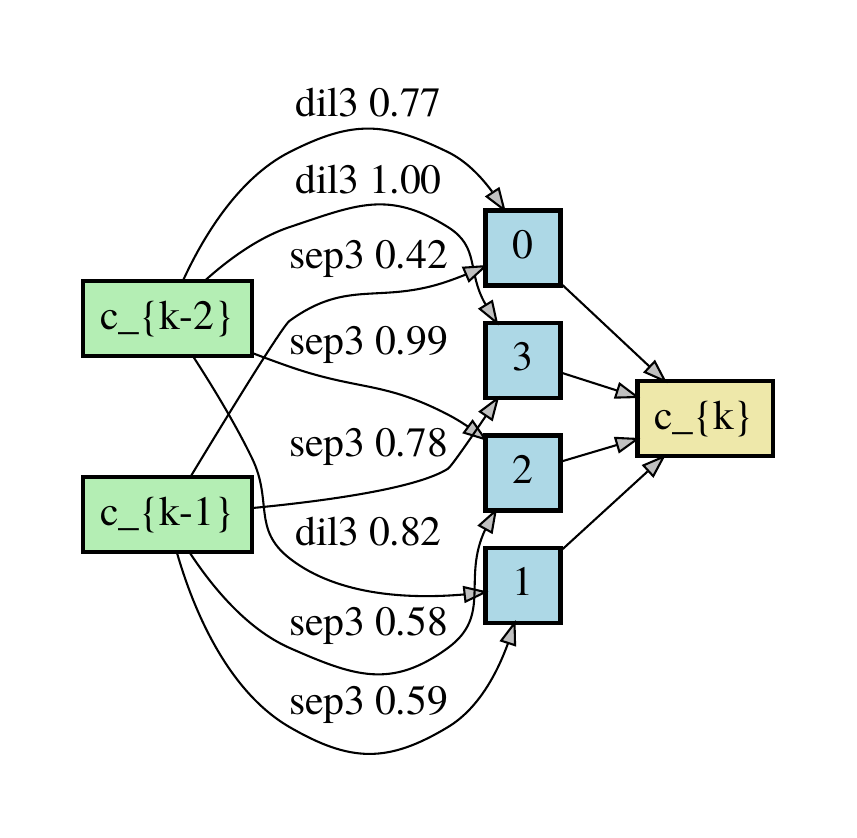}}
\hspace{4em}
\subfigure[The reduction cell on CIFAR-10]
{\label{aca_cifar_rcells}\includegraphics[width=0.41\linewidth]{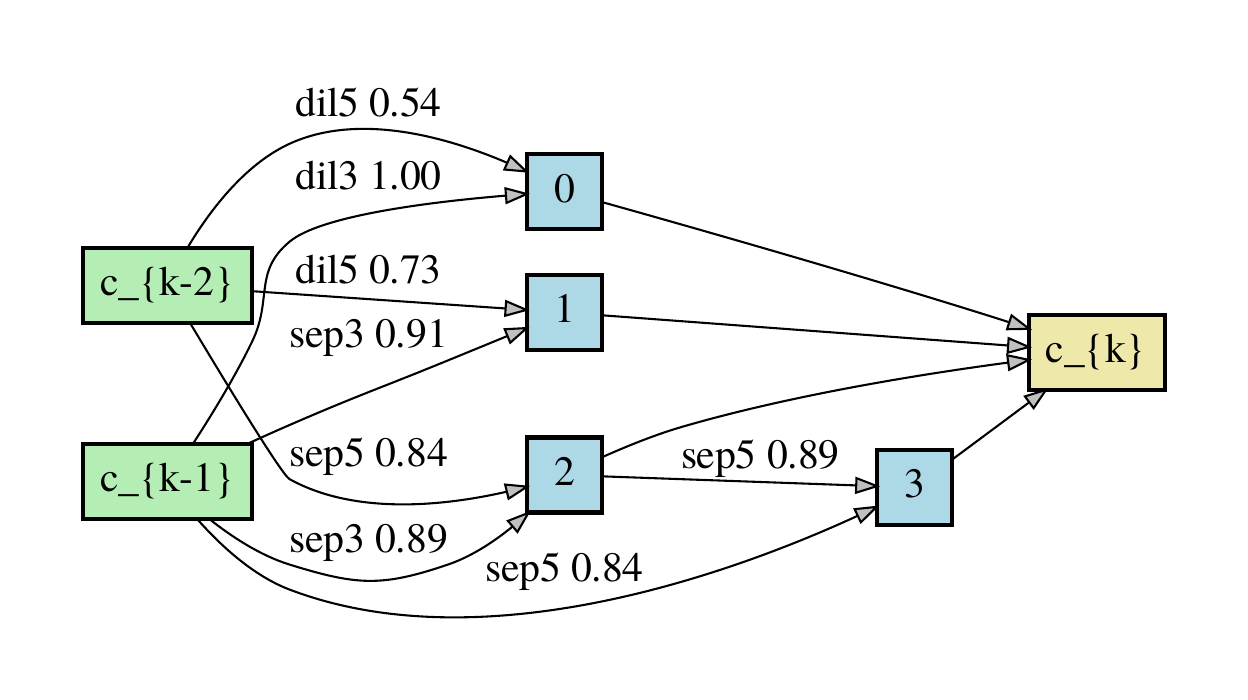}}
\subfigure[The normal cell on ImageNet]
{\label{aca_im_ncells}\includegraphics[width=0.41\linewidth]{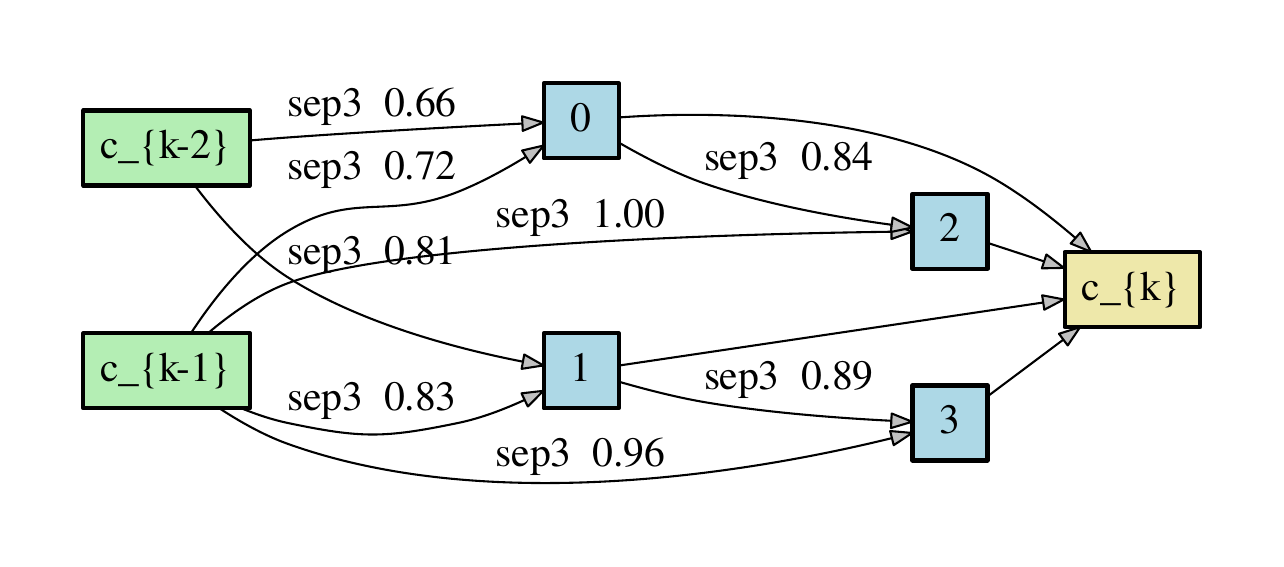}}
\hspace{1em}
\subfigure[The reduction cell on ImageNet]
{\label{aca_im_rcells}\includegraphics[width=0.36\linewidth]{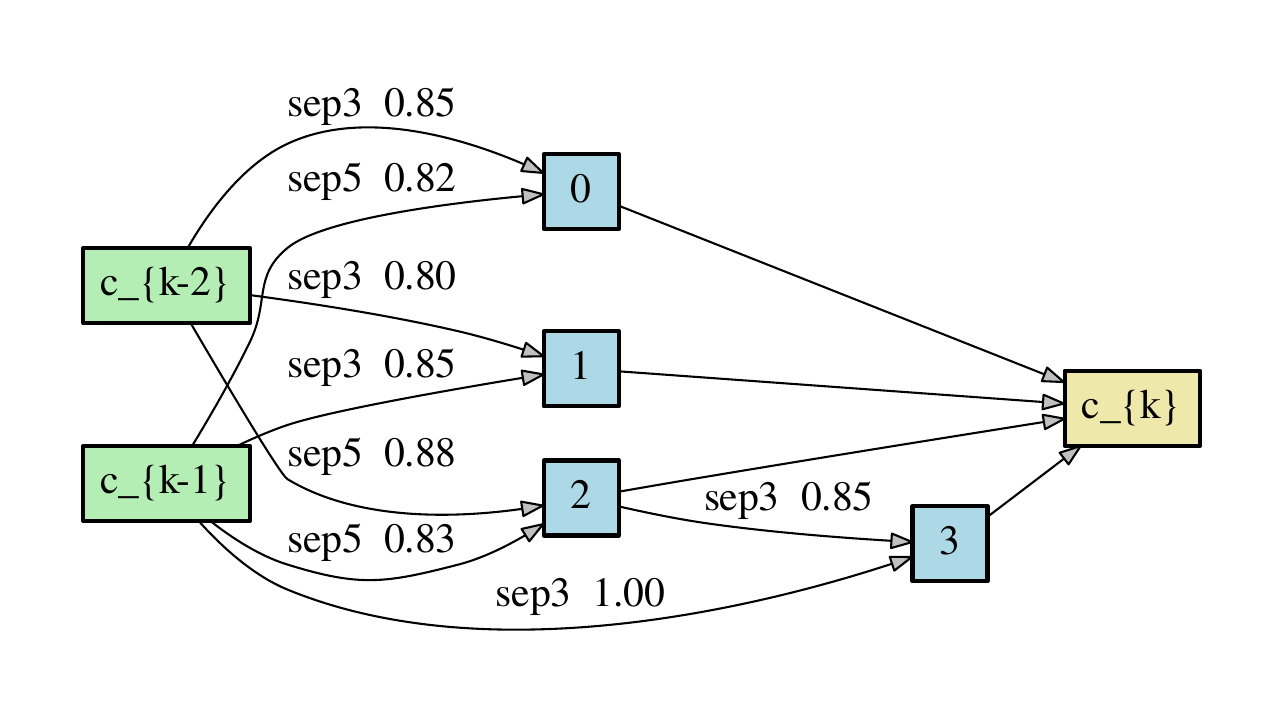}}
\caption{The searched normal cells and reduction cells of ACA-DARTS on CIFAR-10 and ImageNet. The number behind each operation represents its channel ratio. The most important operation has a channel ratio of 1.}
\label{fig:searched_cells_DART_imagenet}
\end{figure*}

%% file: fig/searched_architecture.tex
\begin{figure*}[!t]
\centering
\subfigure[The architecture searched under the complex search space when searching for 50 epochs]
{\label{50_architecture}\includegraphics[width=0.98\linewidth]{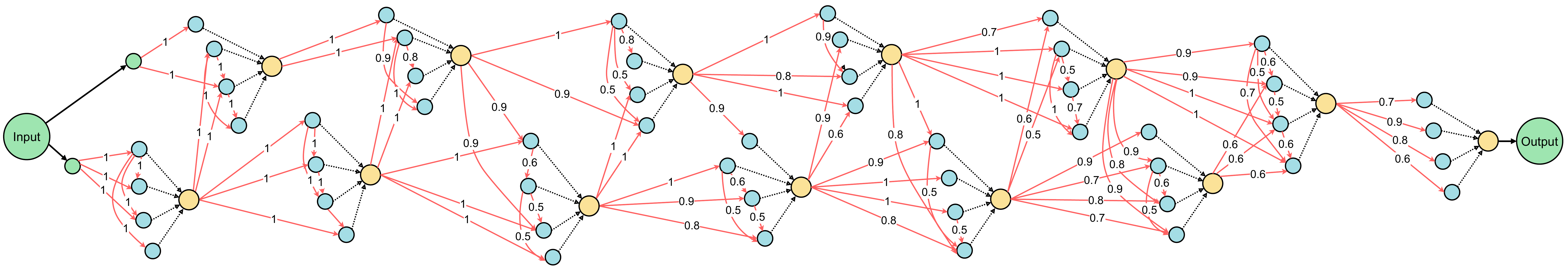}}
\subfigure[The architecture searched under the complex search space when searching for 200 epochs]
{\label{200_architecture}\includegraphics[width=0.98\linewidth]{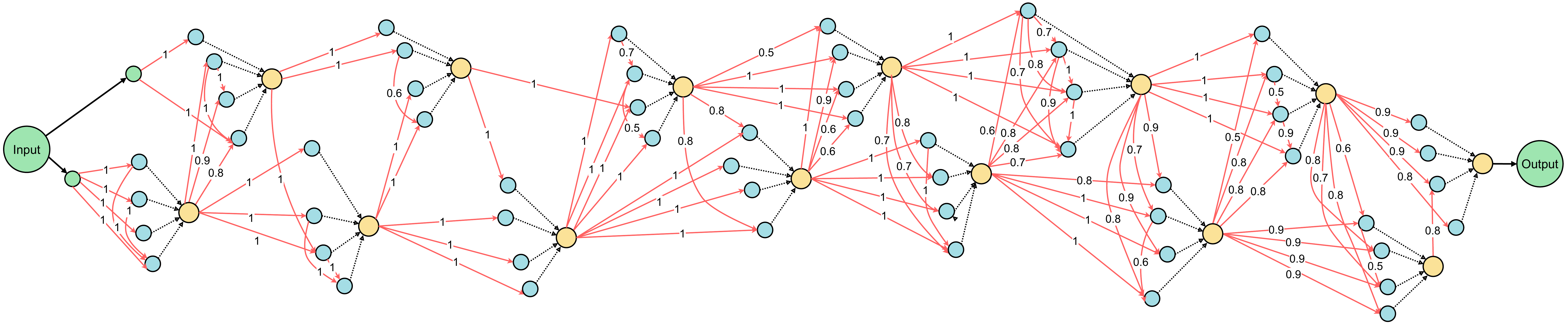}}
\caption{\Revise{Two architectures found on CIFAR-10 under the complex search space. The red thin and black dashed arrows indicate $3\times3$ \emph{SepConv} and concatenation, respectively. The number in each operation represents its channel ratio. Due to space constraints, the channel ratios retain one significant digit. This figure is best viewed in a colored and zoomed-in document.}}
\label{fig:searched_architecture}
\end{figure*}

%% file: fig/n.tex
\begin{figure}[!t]
	\centering
		\includegraphics[width=0.75\linewidth]{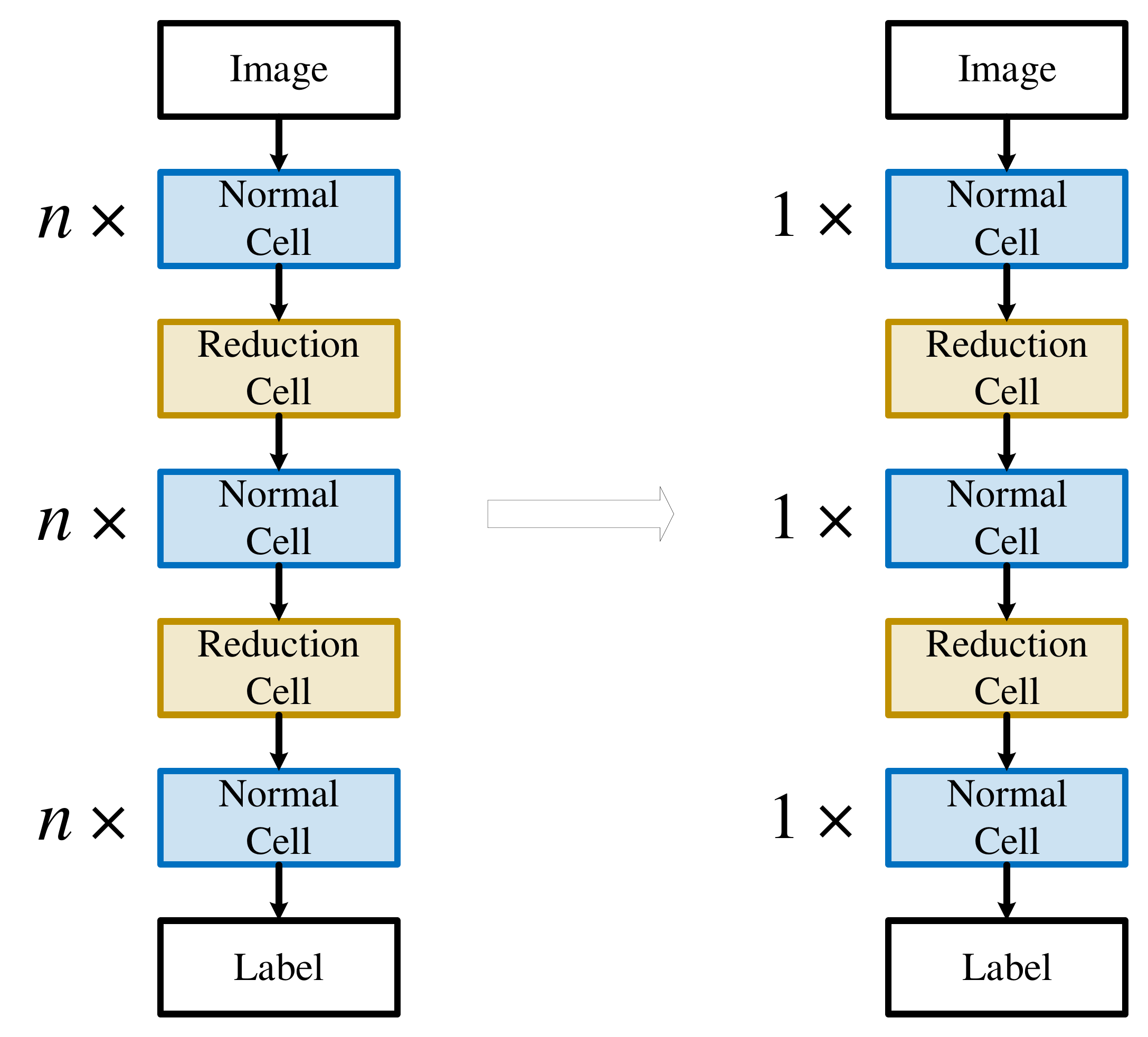}

		\caption{Illustration on the standard macro architecture (\textit{left}) and reduced macro architecture (\textit{right}) in the search stage.}
		\label{fig:n}
\end{figure}

%% file: 05Analysis.tex
\subsection{Exploration the Robustness on Cell Space}
\label{sec:cell_space}
\input{fig/n_results}
In DARTS-based approaches, there are two types of cells in both super-net and target-net. Cells located at the $1/3$ and $2/3$ of the total depth of the network are reduction cells, and the others are normal cells. As shown by the left figure of Fig.~\ref{fig:n}, the super-net of DARTS-based approaches could be divided into $5$ parts: $n$ \emph{normal cells}, \emph{one reduction cell}, $n$ \emph{normal cells}, \emph{one reduction cell}, $n$ \emph{normal cells}. 
So, the depth of the super-net $L=3\times n +2$, where $n$ is the number of repeated normal cells separated by the reduction cell. 
In different DARTS-based approaches, $n$ ranges from $2$ to $5$.

Given that only one cell in each part armed with learnable parameters suffices to enable differentiable search, we aim to narrow the cell space. In the right figure of Fig.~\ref{fig:n}, we reduce the number of normal cells from $n$ to $1$. As shown in Fig.~\ref{fig:cell_space}, DARTS has a dramatic decline at $n=1$ compared to larger $n$, whereas ACA-DARTS has considerably stable performance. So ACA-DARTS is robust to the reduction of cell space. Because the number of normal cells is reduced from $n$ to $1$, the ACA strategy can improve the efficiency of differentiable architecture search by training with smaller computation overhead and larger batches.

\subsection{Exploration the Robustness on Operation Space}
\label{sec:op_space}
The standard DARTS space $\mathbb{S}$ contains 8 operations: 
$\{3\times3$ \emph{SepConv} (sep3), $5\times5$ \emph{SepConv} (sep5), $3\times3$ \emph{dilated SepConv} (dil3), $5\times5$ \emph{dilated SepConv} (dil5), $3\times3$ \emph{MaxPool} (max3), $3\times3$ \emph{AvePool} (avg3), \emph{SkipConnect}, and \emph{Zero}$\}$. 

\input{Table/operation_results}
Zela \etal\cite{zela2019understanding} propose four popular reduced spaces $\mathbb{S}1-\mathbb{S}4$ to demonstrate that the derived architecture in DARTS collapses when \textit{skip connection} dominates the generated architecture. Similarly, we propose another three reduced spaces $\mathbb{S}5-\mathbb{S}7$ to explore the robustness on reduced operation space.
\begin{enumerate}[leftmargin=0.8cm]
    \item[\textbf{$\mathbb{S}5:$}] This search space discards skip connections compared to $\mathbb{S}$. It is the basic search space of ACA-DARTS, corresponding to $\mathbb{S}$ of DARTS-based approaches.
    \item[\textbf{$\mathbb{S}6:$}] In this space, the pooling operations and $Zero$ are further discarded. So the set of candidate operations is $\{3\times3$ \emph{SepConv}, $5\times5$ \emph{SepConv}, $3\times3$ \emph{dilated SepConv} , $5\times5$ \emph{dilated SepConv}$\}$. We discard pooling operations because there are fixed pooling operations at the head and tail of the target-net in the DARTS framework. Besides, the main body of the excellent manual architectures does not contain pooling operations. \emph{Zero} operation is discarded at the end of the search stage in the original DARTS, so we remove it from the operation space.
    \item[\textbf{$\mathbb{S}7:$}] Compared to $\mathbb{S}6$, $\mathbb{S}7$ has one more operation of \emph{SkipConnect}. The set of candidate operations includes $\{3\times3$ \emph{SepConv}, $5\times5$ \emph{SepConv}, $3\times3$ \emph{dilated SepConv}, $5\times5$ \emph{dilated SepConv}, and \emph{SkipConnect}$\}$. This space will be used for DARTS-based approaches as a natural control group.
\end{enumerate}

We investigate the performance of ACA-DARTS in the $\mathbb{S}5$ and $\mathbb{S}6$ operation spaces \Revisegreen{on CIFAR-10}. As a comparison, we also show the results of several DARTS-based approaches in the $\mathbb{S}$ and $\mathbb{S}7$ operation spaces. 
As shown in Table~\ref{tab:op_space}, the performance of DARTS-based approaches drops dramatically after pooling operations and \emph{Zero} be removed from the operation space (from $\mathbb{S}$ to $\mathbb{S}7$). Specifically, DARTS has a decline of $1.29\%$ for DARTS, PC-DARTS has a decline of $0.51\%$, and DARTS$-$ has a decline of  $1.31\%$. In contrast, ACA-DARTS is robust to the absence of pooling operations and \emph{Zero} (from $\mathbb{S}5$ to $\mathbb{S}6$). Hence, we can conclude that ACA-DARTS is far more robust to the absence of relatively unimportant operations than other DARTS-based approaches. Because the number of operations in $\mathbb{S}6$ is half that of $\mathbb{S}$, the ACA strategy can further improve the efficiency of differentiable architecture by employing a smaller operation space without losing any accuracy.

\section{Ablation Study}

\input{Table/ablation}
\subsection{Necessity of the Refilling} 
We conduct ablation studies on refilling skip connections and adaptive channel allocation to better understand their function. 
The results \Revisegreen{on CIFAR-10} are shown in Table~\ref{tab:ablation}. When the target-net does not contain refilled skip connections, the evaluation stage is the same as DARTS, and the performance declines dramatically. The skip connections can not be refilled automatically when the channel allocation strategy is removed, so we randomly allocate eight channels to the skip connections. The performance suffers considerably without channel allocation, verifying that the contribution of different retained operations is not identical. Channel importance ranking is part of adaptive channel allocation. When removing it from adaptive channel allocation by using random allocation, the performance of our method also declines, demonstrating that allocating unimportant channels to skip connections is a better strategy than random allocation. \Revisegreen{
The architectures without refilled skip connections and
without adaptive channel allocation are illustrated in  Figure 1 and Figure 2, respectively, in the Appendix.
}

\input{Table/DARTS_variance}
\subsection{Orthogonal Combination with PC-DARTS}
PC-DARTS employs a partial channel strategy, which is 
comparable to our channel allocation strategy. The main difference is that their channel sampling strategy works on multiple candidate operations of each edge in the search stage to reduce the computational overhead with a hyper-parameter $K$ that requires careful calibration, whereas our channel allocation strategy works on one retained operation of each edge in the target-net to stabilize the searched architecture and achieves operation-wise attention. Another difference is that the channels are randomly sampled in PC-DARTS, while the channels in the channel allocation strategy are filled based on the channel importance. 
For a fair comparison of channel sampling in PC-DARTS with (w/) and without (w/o) ACA strategy, we remove the strong prior of channel shuffling, and then experiments are repeated with different seeds under the same setting. 

As shown in Table~\ref{tab:pcdarts-dart}, PC-DARTS combines ACA strategy (ACA-PC-DARTS) can marginally improve the performance of PC-DARTS, which is $0.19\%$ on average for CIFAR-10 and $0.50\%$ on average for CIFAR-100.

%% file: fig/n_results.tex
\begin{figure}[!t]
	\centering
		\includegraphics[width=0.9 \linewidth]{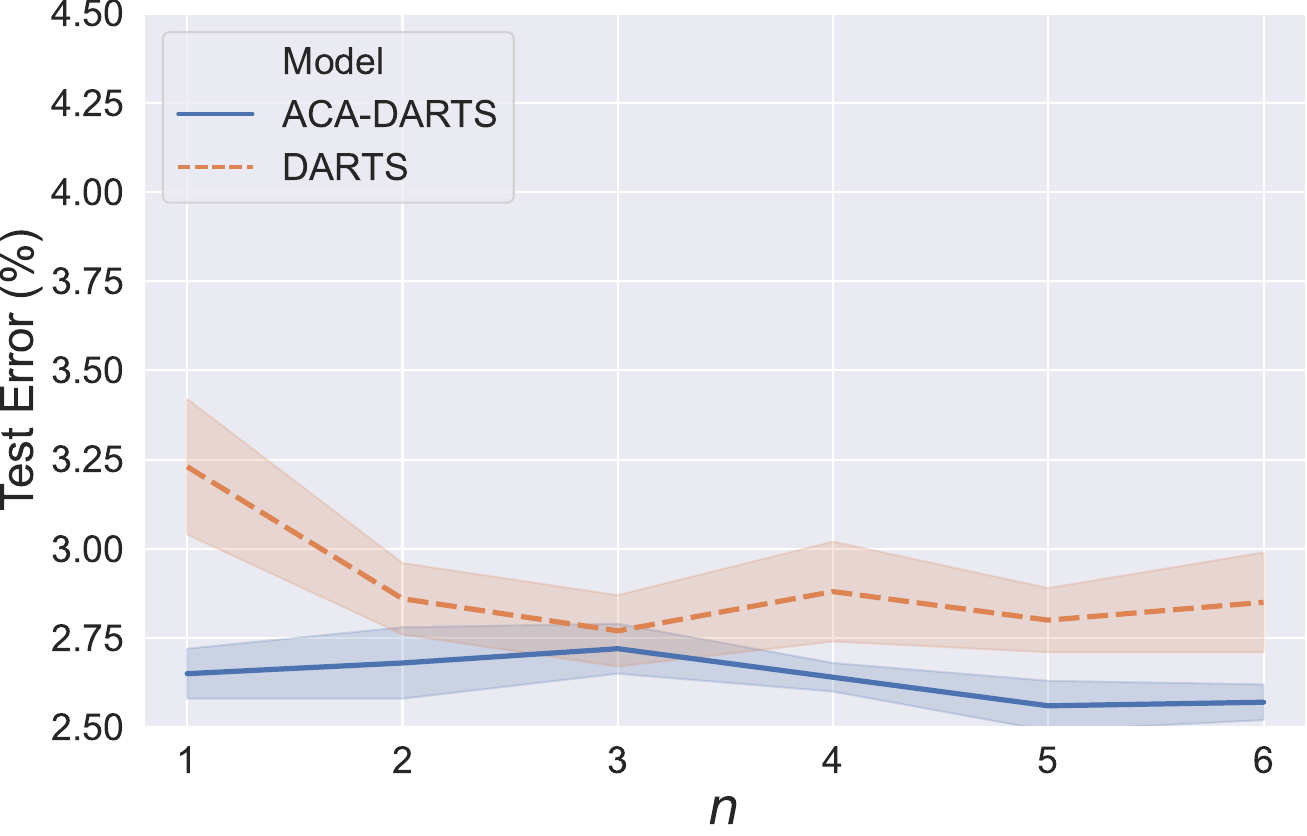}

		\caption{The performance of DARTS (2nd) and our ACA-DARTS (2nd) \Revisegreen{on CIFAR-10} under different number of repeated normal cells $n$.}
		\label{fig:cell_space}
\end{figure}

%% file: Table/operation_results.tex
\begin{table}[!t]
	\centering
	\caption{The performance of DARTS-based approaches and our ACA-DARTS under standard operation space and reduced operation space. $\mathbb{S}5$ and $\mathbb{S}6$ are the skip-connection-free version of $\mathbb{S}$ and $\mathbb{S}7$, respectively.
 } 
	\begin{threeparttable}[b]
	\small
	\resizebox{0.4\textwidth}{!}{
\begin{tabular}{lcc}
					\hline
					\textbf{Method} & \textbf{$\mathbb{S}$ } & \textbf{$\mathbb{S}7$} \\
					\hline
					DARTS (2nd) & 2.76$\pm$0.09 & 4.05$\pm$0.15\\
					PC-DARTS & 2.57$\pm$0.07 & $3.08\pm$0.06\\
					DARTS$-$ & 2.59$\pm$0.08 & 3.90$\pm$0.17 \\
					\hline
					\textbf{Method} & $\mathbb{S}5$ & $\mathbb{S}6$ \\
					\hline
					ACA-DARTS (2nd) & 2.70$\pm$0.08 & 2.68$\pm$0.09 \\
     				ACA-PC-DARTS & 2.56$\pm$0.11 & $2.54\pm$0.09\\
					ACA-DARTS$-$ & 2.59$\pm$0.10 & 2.59$\pm$0.09\\
					\hline
\end{tabular}}
\end{threeparttable}
\label{tab:op_space}
\end{table}

%% file: Table/ablation.tex
\begin{table}[!t]
	\centering
	\caption{Ablation study of refilled skip connections (\textit{w/o skip}), adaptive channel allocation (\textit{w/o channel}) and channel importance ranking (\textit{w/o ranking}) in ACA-DARTS (2nd) \Revisegreen{on CIFAR-10.}} 
	\begin{threeparttable}[b]
	\small
	\resizebox{0.40\textwidth}{!}{
	
	\begin{tabular}{lc}
		\toprule
		\textbf{Method} & \textbf{Test Error (\%)} \\
		\midrule
		ACA-DARTS w/o skip   & 3.38$\pm$0.12\\
		ACA-DARTS w/o channel& 2.86$\pm$0.08 \\
        ACA-DARTS w/o ranking& 2.75$\pm$0.08 \\
		ACA-DARTS     & 2.68$\pm$0.09 \\
		\bottomrule
	\end{tabular}}
	\end{threeparttable}
	\label{tab:ablation}
\end{table}

%% file: Table/DARTS_variance.tex
\begin{table}
\centering
\caption{The test error of PC-DARTS w/ and w/o the ACA strategy with $K=2$ after removing the strong prior of channel shuffling (CS).}
\small
\label{tab:pcdarts-dart}
\begin{threeparttable}[b]
\small
\resizebox{0.48\textwidth}{!}{
\begin{tabular}{lccc}
\toprule
\textbf{Method}  & \textbf{CIFAR-10 } & \textbf{CIFAR-100 } &\textbf{Cost}$^\ddagger$ \\
\midrule
PC-DARTS w/o CS & 2.89$\pm$0.15 &17.14$\pm$0.32 & 0.1 \\
ACA-PC-DARTS w/o CS   & \textbf{2.70$\pm$0.07} &\textbf{16.64$\pm$0.14}    & 0.1 \\
\bottomrule
\end{tabular}}
\begin{tablenotes}
\footnotesize
\item[$\ddagger$] Search cost (GPU-days).
\end{tablenotes}
\end{threeparttable}
\end{table}


